\documentclass[letterpaper]{article} 
\usepackage{aaai2026}  
\usepackage{booktabs}
\usepackage{times}  
\usepackage{algorithm}
\usepackage{algorithmic}
\usepackage{helvet}  
\usepackage{courier}  
\usepackage[hyphens]{url}  
\usepackage{graphicx} 
\usepackage{natbib}  
\usepackage{caption} 
\makeatletter
\@ifundefined{isChecklistMainFile}{
  \newif\ifreproStandalone
  \reproStandalonetrue
}{
  \newif\ifreproStandalone
  \reproStandalonefalse
}
\makeatother

\ifreproStandalone
\usepackage{newfloat}
\usepackage{enumerate}
\usepackage[table]{xcolor}
\usepackage{graphicx}
\usepackage{tabularx}
\usepackage{placeins} 
\usepackage{xcolor}
\usepackage{times}
\usepackage{helvet}
\usepackage{courier}
\DeclareCaptionStyle{ruled}{labelfont=normalfont,labelsep=colon,strut=off} 

\usepackage{listings}
\fi
\makeatletter\def\@listi{\leftmargin\leftmargini \topsep .5em \parsep .5em \itemsep .5em}
\def\@listii{\leftmargin\leftmarginii \labelwidth\leftmarginii \advance\labelwidth-\labelsep \topsep .4em \parsep .4em \itemsep .4em}
\def\@listiii{\leftmargin\leftmarginiii \labelwidth\leftmarginiii \advance\labelwidth-\labelsep \topsep .4em \parsep .4em \itemsep .4em}\makeatother

\newcounter{checksubsection}
\newcounter{checkitem}[checksubsection]

\usepackage{listings}
\usepackage{bibentry}

\usepackage{multirow}
\usepackage{algorithm}
\usepackage{algorithmic}
\usepackage{times}  
\usepackage{helvet}  
\usepackage{courier}  
\usepackage{adjustbox}
\usepackage[hyphens]{url}  
\usepackage{graphicx} 
\usepackage{natbib}  
\usepackage{caption} 
\usepackage{amsmath}
\usepackage{amssymb}

\usepackage{amsmath}
\usepackage{amssymb}
\usepackage{newfloat}
\usepackage{listings}
\DeclareCaptionStyle{ruled}{labelfont=normalfont,labelsep=colon,strut=off} 
\floatstyle{ruled}
\newfloat{listing}{tb}{lst}{}
\floatname{listing}{Listing}
\usepackage{bm}
\usepackage{multirow}
\usepackage{comment}

\title{MotionPhysics: Learnable Motion Distillation for Text-Guided Simulation}
\author{
    Miaowei Wang, 
    Jakub Zadrożny, 
    Oisin Mac Aodha, 
    Amir Vaxman
}
\affiliations{
School of Informatics, The University of Edinburgh, Edinburgh EH8 9AB, United Kingdom

m.wang-123@sms.ed.ac.uk, \{jakub.zadrozny, oisin.macaodha, a.vaxman\}@ed.ac.uk
}

\usepackage{bibentry}

\begin{document}
\twocolumn[{%
\renewcommand\twocolumn[1][]{#1}%
\begin{center}
    \centering
    \maketitle
    \captionsetup{type=figure}
\includegraphics[width=0.95\textwidth, trim=0 0 0 0, clip]{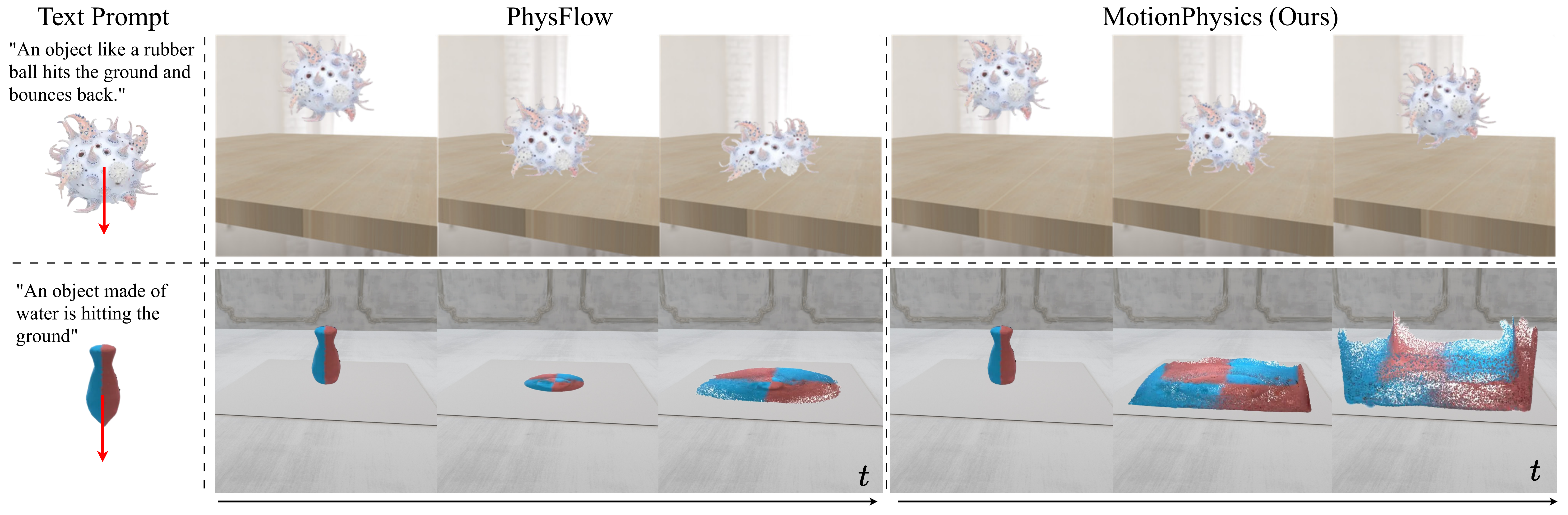}
    \captionof{figure}{
        \textbf{MotionPhysics} automatically estimates plausible material parameters to support dynamic 3D simulation of diverse materials and object types. Compared to prior work (e.g., PhysFlow \cite{liu2025physflow}), it more accurately adheres to the user's input prompt (Left), particularly for AI-generated objects (Top: elastic simulation), human-designed objects (Bottom: water simulation), and real-world scans. Image backgrounds are  from~\citet{li2023pac}.
    }
\label{fig:teaser}
\end{center}
}]

\begin{abstract}

Accurately simulating existing 3D objects and a wide variety of materials often demands expert knowledge and time-consuming physical parameter tuning to achieve the desired dynamic behavior.
We introduce \textit{MotionPhysics}, an end‑to‑end differentiable framework that infers plausible physical parameters from a user-proved natural language prompt for a chosen 3D scene of interest, removing the need for guidance from ground‑truth trajectories or annotated videos. 
Our approach first utilizes a  multimodal large language model to estimate  material parameter values, which are constrained to be within plausible ranges. 
We further propose a learnable motion distillation loss, which extracts robust motion priors from pretrained video diffusion models while minimizing appearance and geometry inductive biases to guide the simulation. We evaluate \emph{MotionPhysics} across more than thirty scenarios, including real-world, human-designed, and AI-generated 3D objects, spanning a wide range of materials such as elastic solids, metals, foams, sand, and both Newtonian and non-Newtonian fluids. We demonstrate that it produces visually realistic dynamic simulations guided by natural language, surpassing the state of the art, with physically plausible parameters that are automatically determined. 

The code and project page are available at:  \url{https://wangmiaowei.github.io/MotionPhysics.github.io/}.
\end{abstract}

\section{Introduction}
The specification of plausible physical parameters is essential for realistic 3D simulation \cite{gottstein2004physical}. For instance, Young’s modulus controls a material’s stiffness, while yield stress marks the onset of irreversible plastic deformation. Traditional methods for identifying such parameters often rely on expert intuition or laborious trial-and-error \cite{snowsimulation}, making simulation pipelines time-consuming and inaccessible to non-experts.

This has motivated a wide body of work focused on automatic physical parameter estimation. Early approaches relied on direct observations \cite{asenov2019vid2param,wu2016physics,jaques2020physics,ma2023learning}. More recent work in novel view synthesis, such as Neural Radiance Fields (NeRF) \cite{mildenhall2021nerf} and Gaussian Splatting (GS) \cite{kerbl20233d}, offers alternative strategies. GS represents scene geometry explicitly via Gaussian kernels, which enables straightforward integration with existing simulators such as PBD for fluids \cite{feng2024splashing}, XPBD for elastic bodies \cite{jiang2024vr}, and MPM for general materials \cite{xie2024physgaussian}. This integration has revived interest in system identification, which recovers physical parameters from multi‑view videos of synthetic objects \cite{gradsim,cai2024gaussian,li2023pac}. When ground‑truth dynamics are unavailable, recent methods guide video diffusion models with text or image prompts to infer plausible parameter values \cite{liu2025physflow,zhang2024physdreamer,lin2025omniphysgs,huang2025dreamphysics}.

While video diffusion models offer the potential for  zero-shot parameter estimation, 
their performance remains limited. Recent studies \cite{phyworld,bansal2024videophy} report that both open-source \cite{yang2024cogvideox,chen2024videocrafter2,wang2025lavie,zheng2024open} and closed-source \cite{pika_art,bar2024lumiere,luma_dream_machine} video generation systems fail to produce videos that obey even basic physical common sense. Consequently, for many objects and scenes such as AI-generated shapes (Fig.~\ref{fig:teaser}, Top), human-designed models (Fig.~\ref{fig:teaser}, Bottom), or real-world objects viewed from novel viewpoints or subjected to different force conditions (Fig.~\ref{fig:robust}), current video diffusion-based methods  often infer incorrect simulation parameters.

Building fully-fledged video diffusion models capable of generating physically plausible outputs is impractical due to their massive computational demands and the need for diverse ground-truth (GT) motion data across a wide range of novel objects. Instead, we leverage existing pretrained video diffusion models to assist with out-of-distribution scene and object simulations. We aim to distill plausible motion cues from pre-trained models guided by high-level, user-provided language instructions, while mitigating the models' inductive shape and appearance biases. To
achieve this, we introduce a novel Learnable Motion Distillation (LMD) loss that extracts pure \emph{motion} signals from a pretrained video diffusion model to steer our differentiable simulations. Concretely, LMD minimizes appearance and geometry discrepancies between the simulation and the diffusion model’s predictions by combining a lightweight, trainable motion extractor with augmented perturbations in both geometry and appearance during training.

Accurate initialization of simulation parameters is critical as poor starting values waste computation and hinder convergence.  We extend PhysFlow’s multimodal initialization with constraint-aware prompts that embed domain-specific parameter bounds (e.g., typical Young’s modulus or density ranges for metal, foam, plasticine). By forcing the LLM to select values within these limits, we leverage its internal knowledge of real-world materials. This approach both anchors simulations in physically plausible ranges and suppresses LLM hallucinations and fabrications~\cite{farquhar2024detecting,hao2024quantifying,walters2023fabrication}.

Our ultimate goal is to ensure that \emph{``what you describe is exactly what you simulate''}. We validate our framework on over 30 simulation scenarios, spanning elastic materials, plasticine, metals, foams, sands, Newtonian, and non‑Newtonian fluids. Our main contributions are:
(i) the introduction of a learnable motion distillation loss to isolate and leverage true motion signals, with LLM initialization triggered by plausible material range values, and 
(ii) a fully automatic, text‑guided  system that achieves state‑of‑the‑art (SOTA) simulation performance, surpassing existing methods seamlessly on human‑designed, AI‑generated, and real‑world objects.

\begin{figure*}[!htb]
\centering
\includegraphics[width=0.95\textwidth]{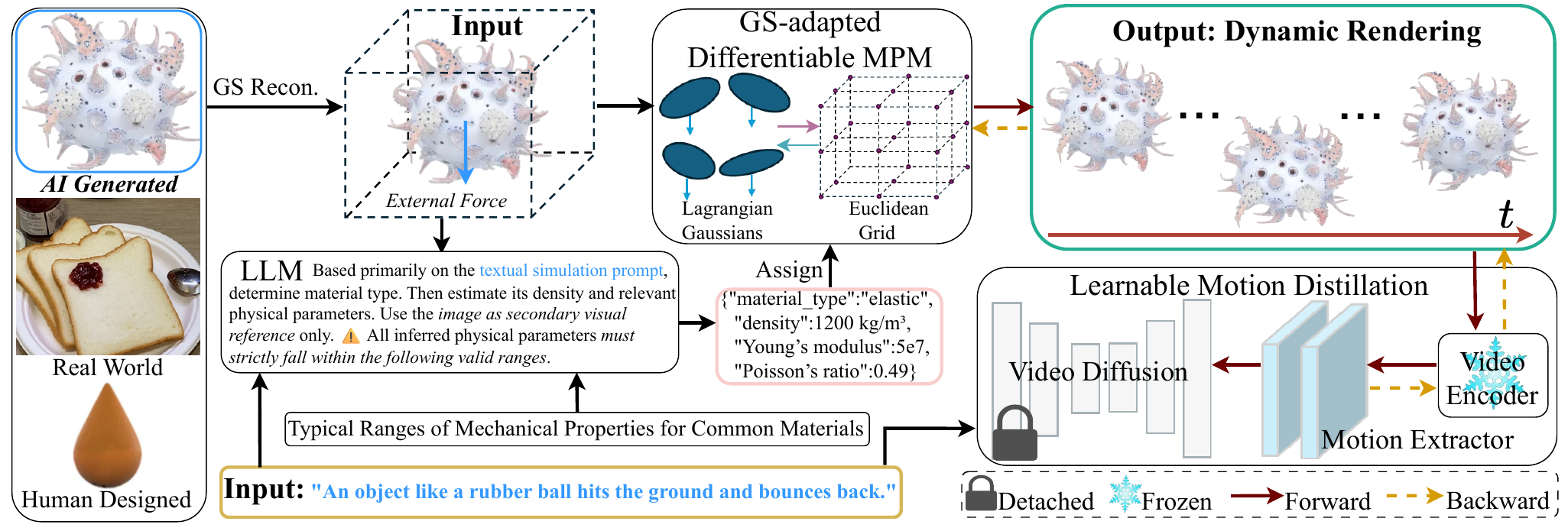}
\caption{\textbf{Overview.} MotionPhysics simulates physically consistent dynamics from text-guided input prompts by automatically estimating physical parameters for diverse input scenes, including AI-generated, real-world, and human-designed assets.
}
\label{fig:pipeline}
\end{figure*}

\section{Related Work}
\noindent \textbf{\emph{3D Dynamics Generation.}}  Diffusion~\cite{liu2022video} and flow-matching~\cite{jin2024pyramidal} models have enabled medium duration, high-quality video synthesis from text or image prompts, exemplified by \textsc{Sora}~~\cite{liu2024sora} and \textsc{Goku}~~\cite{chen2025goku}. 
To extend these dynamic priors into 4D (3D + time), several works fuse diffusion with dual shape–texture representations, using either NeRF-style encoders or Gaussian Splatting (GS)
to generate view-consistent, dynamic scenes without explicit 4D supervision~\cite{pmlr-v202-singer23a,yuan20244dynamic,ling2024align,zhao2023animate124}. 
While implicit methods such as D-NeRF \cite{pumarola2021d} and HyperNeRF~\cite{park2021hypernerf} often suffer from slow rendering and limited user control, explicit splatting pipelines (e.g., \textsc{4DGS} \cite{wu2024_4dgs}, \textsc{Hybrid3D–4DGS} \cite{oh2025hybrid3d4dgaussiansplatting}, and other real-time systems~\cite{yang20244d,duan20244d}) offer fast, editable alternatives. These splatting frameworks have been further generalized across input modalities: multi-view static reconstruction~~\cite{chen2024pgsr,Huang2DGS2024}, dynamic captures~~\cite{xu2024grid4d}, single uncalibrated images~~\cite{yi2023gaussiandreamer,smart2024splatt3r}, and mesh-to-Gaussian-field conversion~~\cite{waczynska2024games}. Despite these advances, most methods lack physics grounding, as they do not model deformations or motion driven by forces and material properties, limiting realism~\cite{bansal2025videophy2,zhang2025morpheus,cao2024physgame,wang2025ns4dynamics}.

\noindent \textbf{\emph{Physics-grounded Dynamic Generation.}}    Embedding physical laws into generative models \cite{zhong2024springgaus} yields more realistic interactions and using differentiable simulators enables gradient‐based motion synthesis while simultaneously tuning material parameters. 
Simple spring–mass systems, such as \textsc{SpringGaus} \cite{zhong2024springgaus} and \textsc{PhysTwin} \cite{jiang2025phystwin}, effectively capture elastic deformation, while the differentiable Material Point Method (MPM)~\cite{jiang2016material} excels at modeling diverse material behaviors. For example, 
\textsc{PhysMotion}~\cite{tan2024physmotion} leverages MPM for single-image dynamics, whereas \textsc{PAC-NeRF}~\cite{li2023pac} fuses NeRF and MPM via particles, trading off efficiency and fidelity in complex scenes. \textsc{PhysGaussian}~\cite{xie2024physgaussian} further boosts visual quality by combining 3DGS with MPM, yet still requires hand-tuned material settings. System identification can recover these settings but requires ground-truth (GT) videos~\cite{cai2024gaussian} or markers \cite{ma2023learning}, limiting scalability. Thus, automatic estimation of material properties without any GT dynamics is an open challenge for real-world applications.

\noindent \textbf{\emph{Physical Parameter Estimation.}} To inject physical realism, recent method \cite{zhang2024physdreamer,liu2024physics3d,huang2025dreamphysics,lin2025omniphysgs,liu2025physflow} leverage video-diffusion priors \cite{blattmann2023stable,meng2024towards} to infer material properties such as elasticity and plasticity. \textsc{PhysDreamer} \cite{zhang2024physdreamer} models elastic behavior in real scenes, and \textsc{Phys3D} \cite{liu2024physics3d} extends this approach to plastic deformations. \textsc{DreamPhysics} and \textsc{PhysFlow} handle a broader range of materials, while 
\textsc{OmniPhysGS} integrates constitutive models into each particle to support heterogeneous interactions.
These methods optimize material parameters by backpropagating through differentiable simulators using either (i) direct perceptual objectives, such as image similarity \cite{zhang2024physdreamer} or optical-flow divergence~\cite{liu2025physflow} between simulated and generated frames, or (ii) score-distillation losses derived from a diffusion model~\cite{huang2025dreamphysics,liu2024physics3d}. However, these methods rely on real-world footage, where objects are often anchored, occluded, or subject to noise \cite{decoupledGaussian}, limiting their applicability compared to the increasingly prevalent human-designed and AI-generated assets. Human-designed objects often lack consistent textures (Fig.\ref{fig:teaser}, Bottom), while AI-generated meshes~\cite{hunyuan3d22025tencent,tochilkin2024triposr}  can exhibit atypical geometry or appearance (Fig.\ref{fig:teaser}, Top), confusing appearance-driven supervision. 
To address these limitations, we introduce a \emph{learnable motion distillation} loss that extracts motion cues from pretrained diffusion models guided by text prompts, while suppressing appearance and geometry biases. While some works \cite{jeong2024vmc,zhai2024motion} extract inter-frame motion for video-based transfer, our approach focuses on text-guided estimation of physical parameters for physics-based simulation. 
Text prompts often implicitly convey physical properties by indicating material types and object categories. We enable more effective automated estimation of these material properties by using pretrained LLMs with  plausible material value ranges, a crucial aspect overlooked by prior works~\cite{huang2025dreamphysics,liu2025physflow,lin2024phys4dgen}.

\section{Methodology}
\subsection{Problem Statement}
We consider objects and scenes that are real, synthetic, human-designed (often lacking high-quality textures), or AI-generated (often with uncommon geometry or appearance). These inputs may be provided as multi-view static images, dynamic videos, single images, or meshes. Compared to traditional mesh representations, 3DGS is more suitable for reproducing real-world scenes and supports high-quality, real-time rendering. Methods such as PGSR, GIC \cite{cai2024gaussian}, Splat3R, and GaMes can convert inputs into collections of 3D Gaussians, $
\mathcal{G} := \{\mathbf{x}_g, \sigma_g, \boldsymbol{\Sigma}_g, \mathcal{S}_g\}
$,  where each splat $g$ is defined by its center $\mathbf{x}_g \in \mathbb{R}^3$, opacity $\sigma_g \in [0,1]$, covariance $\boldsymbol{\Sigma}_g \in \mathbb{R}^{3\times3}$, and color coefficients $\mathcal{S}_g$.

To simulate dynamic behavior over discrete time $t$, we denote time-varying splats as $\mathcal{G}^t := \{\mathbf{x}_g^t, \sigma_g^t, \boldsymbol{\Sigma}_g^t, \mathcal{S}_g^t\}$. We adopt a differentiable GS-adapted MLS-MPM simulator \cite{xie2024physgaussian}, which evolves splat states using a Markovian \cite{richey2010evolution} update $\mathcal{T}$, consisting of particle-to-grid and grid-to-particle mappings (velocity omitted for clarity):
\begin{equation}
(\mathbf{x}_g^{\,t+1},\; \boldsymbol\Sigma_g^{t+1},\; \mathcal{S}_g^{t+1})
= 
\mathcal{T}(\mathbf{x}_g^{\,t},\; \boldsymbol\Sigma_g^t,\; \mathcal{S}_g^t 
\mid \bm{\theta},\; \mathbf{f}^{\rm ext}),
\end{equation}
where $\boldsymbol{\Sigma}_g^t$ and $\mathcal{S}_g^t$ evolve through the deformation gradient $F^{t+1}$, which captures stretching, rotation, and shear \cite{xie2024physgaussian}, and $\mathbf{f}^{\rm ext}$ denotes external forces. Following \textsc{PhysFlow} \cite{liu2025physflow}, all splats share the same material parameters defined as $\bm{\theta} := \{\rho, c, \bm{\theta}_c\}$, where $\rho$ is density, $c$ is a material class (e.g., elastic, plasticine, metal, foam, sand, Newtonian or non-Newtonian fluid) corresponding to different material constitutive models, and $\bm{\theta}_c$ contains associated class-specific coefficients (see Suppl.~for details).

\vspace{1mm}
\noindent\textbf{\emph{\underline{Objective:}}} ~Our goal is to \emph{automatically infer} the full material parameter set  $\bm{\theta}$ from a natural language prompt $\mathcal{P}_{\text{text}}$, without supervision from ground-truth dynamics, motion capture markers, or videos. Once obtained, the parameters enable high-fidelity and diverse 3D dynamic simulations under various force fields and physical conditions (see Suppl.).

\subsection{Preliminaries}
Score Distillation Sampling (SDS) \cite{pooledreamfusion} was initially proposed to distill 3D priors from large-scale 2D diffusion models for text-to-3D generation \cite{lin2023magic3d}. Recent extensions \cite{lin2025omniphysgs,huang2025dreamphysics,liu2024physics3d} adapt SDS to optimize physical parameters $\bm{\theta}_c$ using a diffusion model $\phi$. Let $z_0 = \mathcal{E}_\phi(\{\mathcal{I}_l\})$ be the latent encoding of frames $\{\mathcal{I}_l\}$ via the video encoder $\mathcal{E}_\phi$. At video diffusion step $k$, $z_0$ is perturbed by noise $\epsilon \sim \mathcal{N}(0, I)$:
\begin{equation*}
z_k = \sqrt{\bar\alpha_k}\,z_0 \;+\; \sqrt{1 - \bar\alpha_k}\,\epsilon,\quad
\bar\alpha_k = \prod_{i=1}^k \alpha_i,
\quad
\alpha_i \in [0,1].
\end{equation*}
The SDS update is computed as:
\begin{equation}
s_{\mathrm{SDS}}(k, \epsilon) \;=\; w_k \,\Bigl(\epsilon_\phi(z_k, k \mid \mathcal{P}_{\text{text}})\;-\;\epsilon\Bigr),
\end{equation}
where $\epsilon_\phi$ is the predicted noise from $\phi$ and $w_k$ is a step-dependent weight. The gradient with respect to $\bm{\theta}_c$  is:
\begin{equation}
\label{eq:sds}
\nabla_{\bm{\theta}_c} L_{\mathrm{SDS}} 
\;=\; 
\mathbb{E}_{k,\epsilon} \Bigl[\,s_{\mathrm{SDS}}(k, \epsilon)\,\frac{\partial \{\mathcal{I}_l\}}{\partial \{\mathcal{G}^t\}}\;\frac{\partial \{\mathcal{G}^t\}}{\partial \bm{\theta}_c \bf{}}\Bigr].
\end{equation}
Here, $\{\mathcal{I}_l\}$ is differentiably 
rendered from $\{\mathcal{G}^t\}$ by 3DGS,  and $\{\mathcal{G}^t\}$ depends on $\bm{\theta}_c$ via GS-adapted MLS‑MPM simulation. This objective aligns simulated videos with the data distribution of the diffusion model, thereby distilling physical priors from the latter to
optimize the parameters $\bm{\theta}_c$.

\subsection{Method Overview}
As shown in Fig.~\ref{fig:pipeline}, we first reconstruct the static object or scene into an initial GS representation $\mathcal{G}^0$. Simultaneously, 
we prompt a multimodal LLM (e.g., GPT‑4~\cite{achiam2023gpt}) with prescribed parameter ranges defined for each possible material type, conditioning primarily on the user’s text prompt $\mathcal{P}_{\text{text}}$ and secondarily on a reference rendered image, to obtain an initial material parameters estimate $\bm{\theta}^{\rm ini}$. Using $\bm{\theta}^{\rm ini}$ and external forces $\mathbf{f}^{\rm ext}$, our differentiable GS‑adapted MLS‑MPM solver simulates splat dynamics over time,  yielding a sequence $\{\mathcal{G}^t\}$. We alpha‑blend a sparse subset of these into frames $\{\mathcal{I}_l\}$. To refine $\bm{\theta}^{\rm ini}$, we introduce a learnable motion distillation loss (Eq.~\ref{eq:lmd}), which extracts dynamic priors from a pretrained video diffusion model $\phi$ conditioned on the same $\mathcal{P}_{\text{text}}$. By iterating simulation, rendering, and gradient‑based optimization, we converge to a physically plausible parameter set $\bm{\theta}_c$ that faithfully reproduces the motion specified by the user.

\subsection{Initialization via Multimodal LLMs}
\label{sec:LLM}
Accurately identifying the material type $c$ is crucial for selecting the appropriate constitutive model in MLS‑MPM. Given a user prompt, e.g.,  \textit{``A rubber ball-like object hits the ground and bounces back.''}, we aim to infer the material type $c$, density $\rho$, and a corresponding set of material-specific parameters $\bm{\theta}_c$. If the prompt changes, the inferred material and parameters should adapt accordingly, enabling flexible, user-driven simulation (see Figure~A4 in Suppl.).

To achieve this, we leverage GPT-4~\cite{achiam2023gpt} to estimate initial material parameters $\bm{\theta}^{\rm ini}$ primarily from text and secondarily from a reference image.  However, naïvely querying GPT-4 as done by PhysFlow can lead to hallucinated \cite{farquhar2024detecting,hao2024quantifying} or fabricated \cite{walters2023fabrication} numerical values (see Fig.~\ref{fig:ablation_init}). LLMs inherently encode extensive real‑world material knowledge, so to reduce hallucinations, we provide GPT‑4 with prompt templates that contain value‐range constraints grounded in standard material‑property handbooks~\cite{callister_materials,ashby_materials,mitchell_soil,gibson_ashby_foam,crc_handbook}. This grounding steers the LLM toward realistic values, preventing implausible predictions and simulation failures in certain cases (Fig.~\ref{fig:ablation_init}). The Supplement includes the full prompt template and typical parameter ranges. For instance, Young’s modulus spans $10^7 –4\times10^{11}$ Pa for elastic materials and $10^6-5\times10^6$ Pa for plasticine.

\subsection{Learnable Motion Distillation}
\label{sec:lmd}
Density $\rho$ and material class $c$ can be reliably inferred using our LLM-based approach above. However, the class‑specific coefficients $\bm{\theta}_c$ are coarse approximations that are insufficient for precise simulation and require additional supervision, since LLMs lack dynamic modelling and simulation capabilities. Following prior work \cite{liu2024physics3d,huang2025dreamphysics}, we employ a video diffusion model $\phi$ to supervise motion optimization. However, diffusion‑based predictions inherently entangle motion with appearance and geometric biases from their training data, making it challenging to extract pure motion signals, especially for human‑designed or AI‑generated objects whose appearance or structure falls outside the training distribution (see Fig.~\ref{fig:ablation_lmd}).

\begin{figure}[!t]
\centering
\includegraphics[width=0.95\linewidth]{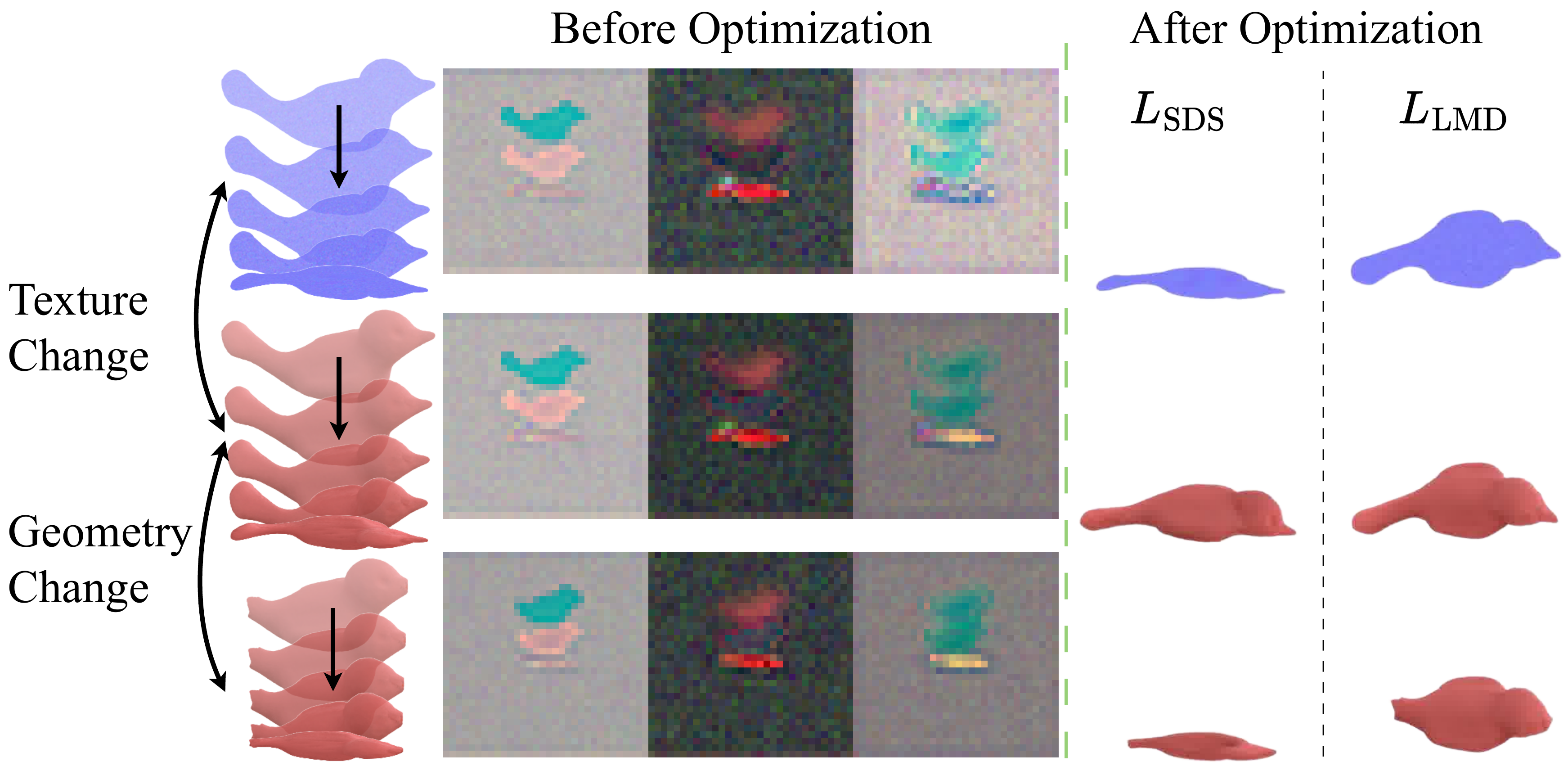}
\caption{\textbf{Structure Similarity.} Left: the same motion pattern (with PCA‑visualized latent codes in the middle) under varying textures and geometries, before optimization. Right: after applying our $L_{\mathrm{LMD}}$, the dynamics become consistent and remain largely unaffected by both texture and geometry.}
\label{fig:latent_visualization}
\end{figure}
\begin{figure*}[!h]
\centering
\includegraphics[width=0.95\textwidth]{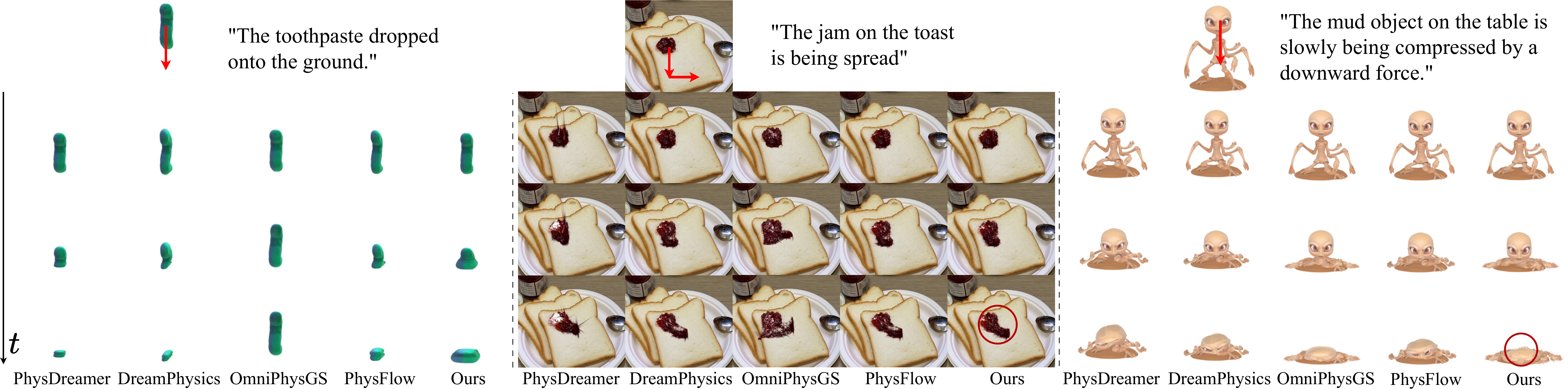}
\caption{\textbf{Qualitative Evaluation.} We compare our method against several baselines across diverse simulation cases, including human-designed objects (e.g., Toothpaste, Left), real-world scenes (e.g., Jam, Middle), and AI-generated objects (e.g., Alien, Right). Red arrows denote input forces, and red circles highlight key regions of difference. See  Suppl. for additional results.
}
\label{fig:qualitative}
\end{figure*}
A key observation is that, despite changes in appearance or shape, identical simulated motions under the same initial physical parameters yield globally consistent latent codes from the video encoder $\mathcal{E}_\phi$. As Fig.~\ref{fig:latent_visualization} (Left) shows, whether the bird’s color shifts from red to blue (Top) or its shape loses tail and beak (Bottom), the latents (visualized by PCA projection) share the same overall structure and vary only in local details. 
This holds across different diffusion models (see Suppl.) and motivates extracting motion signals directly from ``clean" latents rather than noise via a learnable motion extractor $M$, by dynamically smoothing local latent-space disparities, i.e, the appearance and geometry gaps between the model’s pre-learned distribution and the target scene.

Concretely, to enforce motion learning, we augment the initial representation $\mathcal{G}^0$ (both kernel centers and color coefficients) with additive Gaussian noise $\epsilon'$ (see Suppl.), yielding $\tilde{\mathcal{G}}^0$. This produces augmented renderings $\{\tilde{\mathcal{I}}_l\}$ and latents $\tilde{z}_0 = \mathcal{E}_\phi(\{\tilde{\mathcal{I}}_l\})$. Then we compute one‑step denoised latents
 $
\tilde{z}_k^0 = \frac{1}{\sqrt{\bar\alpha_k}}\Bigl(\tilde{z}_k - \sqrt{1-\bar\alpha_k}\,\epsilon_\phi(\tilde{z}_k, k \mid \mathcal{P}_{\text{text}})\Bigr),
$
geometrically projecting $\tilde{z}_k$ back onto the clean‑latent manifold. 

We denote the distilled motion targets and predictions as:
\begin{equation}
y_{\mathrm{target}} = M(\tilde{z}_0),\quad
y_{\mathrm{pred}}   = M(\tilde{z}_k^0),
\end{equation}
where the learnable motion extractor $M$ is a lightweight two-layer convolutional network initialized to the identity mapping and trained with a small learning rate of $2\times10^{-5}$. 
Our resulting learnable motion distillation (LMD) loss uses a Charbonnier variant for numerical stability \cite{barron2019general}:
\begin{equation}
\label{eq:lmd}
L_{\mathrm{LMD}} = w_k\,\sqrt{\|y_{\mathrm{pred}} - y_{\mathrm{target}}\|^2 + \beta^2},
\end{equation}
with small constant $\beta = 10^{-3}$, and $M$ is kept synchronized via exponential moving averaging~\cite{tarvainen2017mean}. Finally, gradients to the coefficients  $\bm{\theta}_c$ are estimated: 
\begin{equation}
\nabla_{\bm{\theta}_c} L_{\mathrm{LMD}}
= \mathbb{E}_{k,\epsilon,\epsilon'}\Bigl[\frac{\partial L_{\mathrm{LMD}}}{\partial \{\tilde{\mathcal{I}}_l\}}
\,\frac{\partial \{\tilde{\mathcal{I}}_l\}}{\partial \{\tilde{\mathcal{G}}^t\}}
\,\frac{\partial \{\tilde{\mathcal{G}}^t\}}{\partial \bm{\theta}_c}\Bigr].
\end{equation}
As shown in Fig.~\ref{fig:latent_visualization} (Right), our $L_{\mathrm{LMD}}$ loss captures consistent motion patterns across appearance and geometry, while $L_{\mathrm{SDS}}$ (Eq.~\ref{eq:sds}) leads to inconsistent results after optimization. 

\section{Implementation Details} 
We build our differentiable simulator on NVIDIA’s WARP implementation \cite{warp2022}. Following \cite{xie2024physgaussian}, we mitigate skinny artifacts via anisotropic regularization and fill solid objects’ internal volumes to enhance simulation realism. To stabilize gradient propagation and improve training speed over long MPM rollouts, we leverage a frame boosting scheme \cite{huang2025dreamphysics,zhang2024physdreamer}. 
Given $M\times L$ total frames (with $M=8$), we split them into $M$ interleaved subsequences $V_i = \{I_i, I_{i+M}, \dots, I_{i+M(L-1)}\}$ for $i = 1, \dots, M$, and alternate supervision across these $M$ groups. Each simulation spans 5 seconds, generating 150 rendered frames in total, with 256 internal substeps per frame. With frame boosting, for each subsequence, we perform $256 \times M$ intermediate updates between adjacent frames, computing gradients only at the final step. Besides, learnable motion distillation is distilled using the CogVideoX model \cite{yang2024cogvideox} with classifier-free guidance (CFG=100), following PhysFlow. Training converges in approximately 40 iterations, with each forward–backward pass taking about 28 seconds on an NVIDIA A100 80 GB GPU. Our framework supports diverse manual specifications of boundary conditions (e.g., see Fig.~\ref{fig:teaser}) and force applications (see Fig.~\ref{fig:robust}), enabling precise spatiotemporal control of material response.

\section{Results and Discussion}
\label{sec:Results}
\subsection{Evaluation Settings}
\noindent\textbf{\emph{\underline{Datasets}.}} We conduct experiments on three  dataset types. 1) \textbf{Human Designed:} We evaluate eight PAC-NeRF models (Torus, Bird, Playdoh, Cat, Trophy, Droplet, Letter Cream, and Toothpaste), which exhibit uniform colors rather than detailed textures. We use the static 3DGS reconstructions from GIC \cite{cai2024gaussian}. Since our focus is text-guided physical simulation using various material prompts (see Suppl.), rather than system identification, we do not use their rendered dynamic frames as ground-truth labels. Those frames rely on manually specified parameters and cannot capture the full diversity of realistic distributions. \\
2) \textbf{Real World:} We include four PhysDreamer scenes (Alocasia, Carnation, Hat, and Telephone) \cite{zhang2024physdreamer} and four additional scenes: Fox from InstantNGP \cite{mueller2022instant}, Plane from NeRFStudio \cite{nerfstudio}, Kitchen from Mip‑NeRF 360 \cite{Barron_2022_CVPR}, and Jam and Sandcastle from PhysFlow. We use the text prompts provided by PhysFlow for all real-world scenes.\\ 
3) \textbf{AI Generated:} We use the meshes Urchin, Alien, Gentleman, and Axe from Hunyuan3D \cite{hunyuan3d22025tencent}, which are further processed with GaMes \cite{waczynska2024games} to obtain their corresponding 3DGS representations.

\vspace{1mm}
\noindent\textbf{\emph{\underline{Baselines}.}} We compare with: 1) \textit{PhysDreamer} \cite{zhang2024physdreamer}, which estimates material properties and initial velocity via an image appearance loss between simulated renderings and generated videos, supporting only elastic materials. 2) \textit{DreamPhysics} \cite{huang2025dreamphysics}, which applies Score Distillation Sampling (SDS). 3) \textit{PhysFlow} \cite{liu2025physflow}, which minimizes an optical flow loss between simulated and generated videos. 4) \textit{OmniPhysGS} \cite{lin2025omniphysgs}, which models heterogeneous objects with constitutive 3D Gaussians guided by video diffusion. We use open-source implementations, where the first three are based on NVIDIA WARP, and OmniPhysGS relies on a slower, less stable PyTorch simulator (see Tab.~\ref{tab:quant-metrics}), so we shorten its video lengths for consistent frame rates. All baselines share identical simulation settings (boundary conditions, external forces, and text prompts). For material initialization, we follow PhysFlow using GPT-4 predictions, except OmniPhysGS, which directly optimizes constitutive models without specific material parameters. All initializations and optimizations, including our method, are run once to ensure fairness.

\vspace{1mm}
\noindent\textbf{\emph{\underline{Metrics}.}}  
We conduct a two-alternative forced-choice (2AFC) user study~\cite{macmillan2005detection,zhang2024physdreamer}, in which 79 participants compare 15 randomly selected side-by-side video pairs, ours versus a baseline under the same scene, prompt, and applied force, and answer: Q1) \textbf{Physical Realism}: \textit{Which video demonstrates a more realistic physical response to the applied force?} and Q2) \textbf{Prompt Adherence}: \textit{Which video demonstrates better adherence to the user prompt?} (see Suppl.~for details). We separately quantify prompt adherence objectively via Overall Consistency (OC) from VBench~\cite{huang2024vbench}, measured with ViCLIP~\cite{wang2023internvid} to capture global semantic alignment between prompt and video, CLIPSIM~\cite{wu2021godiva} (as in OmniPhysGS), averaging the cosine similarity between CLIP~\cite{radford2021learning} embeddings of the prompt and each video frame, and we  assess motion realism using the Energy-Constrained Motion Score (ECMS) from PhysFlow, grounded in the energy‐minimization principle. 

\subsection{Results}
To assess generalization, we varied input forces across diverse scenes and settings, resulting in notable deformations and dynamic responses (see Suppl.).
Qualitative examples are shown in Fig.~\ref{fig:qualitative}, and average user preferences over each baseline are reported in Tab.~\ref{tab:user-study-results-all} (detailed votes in Suppl.). Our method outperforms all baselines in physical realism and prompt adherence, with preferences exceeding 50\% across all datasets and over 80\% on human-designed and AI-generated scenes, demonstrating strong generalization to novel geometries and textures. For example, in the Jam (Fig.~\ref{fig:qualitative}, Middle), our method captures Newtonian viscosity with a smooth, cavity-free spread, unlike the baselines (red circle). Similarly, in Toothpaste (Fig.~\ref{fig:qualitative}, Left), it reproduces non-Newtonian behavior, initially flowing and spreading before settling into a stable mound, while others fail.
\begin{table}[b]
  \centering
  \resizebox{\linewidth}{!}{
    \begin{tabular}{llccc}
      \toprule

      & \textbf{Comparison} & Human Designed $\uparrow$ & Real World $\uparrow$ & AI Generated $\uparrow$\\
      \midrule       
      \multirow{4}{*}{\rotatebox[origin=c]{90}{\footnotesize\textbf{Realism}}}
      
      & Ours vs.\ PhysDreamer      & \textbf{96.77\%} & 77.63\% & \textbf{93.80\%}\\
      & Ours vs.\ DreamPhysics   & \textbf{80.27\%} & 69.77\% & \textbf{94.65\%}\\
      & Ours vs.\ OmniPhysGS     & \textbf{97.62\%} & \textbf{92.73\%} & \textbf{84.60\%}\\
      & Ours vs.\ PhysFlow       & \textbf{90.30\%} & 66.13\% & \textbf{84.20\%}\\
      \midrule
      
      
      \multirow{4}{*}{\rotatebox[origin=c]{90}{\footnotesize\textbf{Adherence}}}
      & Ours vs.\ PhysDreamer      & \textbf{95.88\%} & \textbf{86.08\%} & \textbf{96.45\%}\\
      & Ours vs.\ DreamPhysics   & \textbf{82.35\%} & 75.05\% & \textbf{93.80\%}\\
      & Ours vs.\ OmniPhysGS     & \textbf{96.48\%} & \textbf{91.43\%} & \textbf{80.75\%}\\
      & Ours vs.\ PhysFlow       & \textbf{91.30\%} & 71.53\% & \textbf{85.60\%}\\
      \bottomrule
    \end{tabular}
  }
\caption{\textbf{User Study Results.} Mean percentage values over 80\% (highlighted in bold) 
  show strong preference for physical realism (Top) and prompt adherence (Bottom).}
  \label{tab:user-study-results-all}
\end{table}

\begin{table}[h]
  \centering
  \resizebox{\linewidth}{!}{
    \begin{tabular}{lccccc}
      \toprule
      \textbf{Metrics} & PhysDreamer & DreamPhysics & OmniPhysGS & PhysFlow & \textbf{Ours} \\
      \midrule
      OC $\times 10^{-2} \uparrow$ & 17.00 & 18.02 & 17.10
 & 17.96
 & \textbf{18.18} \\
CLIPSIM $\times 10^{-2} \uparrow$ & 21.62 & 21.64 & 21.27 & 21.32 & \textbf{21.69} \\
      ECMS $\downarrow$      & 27.48 & 15.76 & 13.70 & 13.07 & \textbf{11.37} \\
      \midrule
      Opt.\ Time $\downarrow$ 
      & {16.48\,min} 
      & {18.20\,min} 
      & {$\sim9$\,h}
      & {18.14\,min} 
      & {18.39\,min} \\
    \bottomrule
    \end{tabular}
  }
\caption{
  \textbf{Quantitative Evaluation.} Best results in \textbf{bold}; 
  ``Opt.\ Time'' is optimization time (min/h) after initialization.
}
\label{tab:quant-metrics}
\end{table}
In Tab.~\ref{tab:quant-metrics}, our method achieves the highest average scores across all objective quantitative metrics while maintaining competitive optimization speed (tested on Bird scene from Fig.~\ref{fig:ablation_lmd}, Bottom), thanks to its lightweight motion extractor. However, these metrics do not fully align with human perception. For example, the Droplet scene (see Suppl.) scores slightly lower in CLIPSIM and ECMS than some baselines, despite producing the expected water‑splashing behavior. We attribute this gap to two factors: 1) current video‑text consistency and motion metrics cannot distinguish dynamic differences across identical scenes, forces, and prompts, and 2) pretrained metric models focus on static appearance and lack the ability to capture diverse materials and motion patterns specified by text (see Suppl.~for detailed analysis). One remedy is to compare physical parameters as system identification. However, obtaining GT distributions of physically plausible outcomes is challenging. Similar limitations have been noted in 3D‑generation tasks~\cite{Yu_2025_CVPR,tang2024lgm}. Therefore, we emphasize visual comparisons and present qualitative evaluations in subsequent experiments. 

\subsection{Ablation Study}
\begin{figure}[!htb]
\centering
\includegraphics[width=0.95\linewidth]{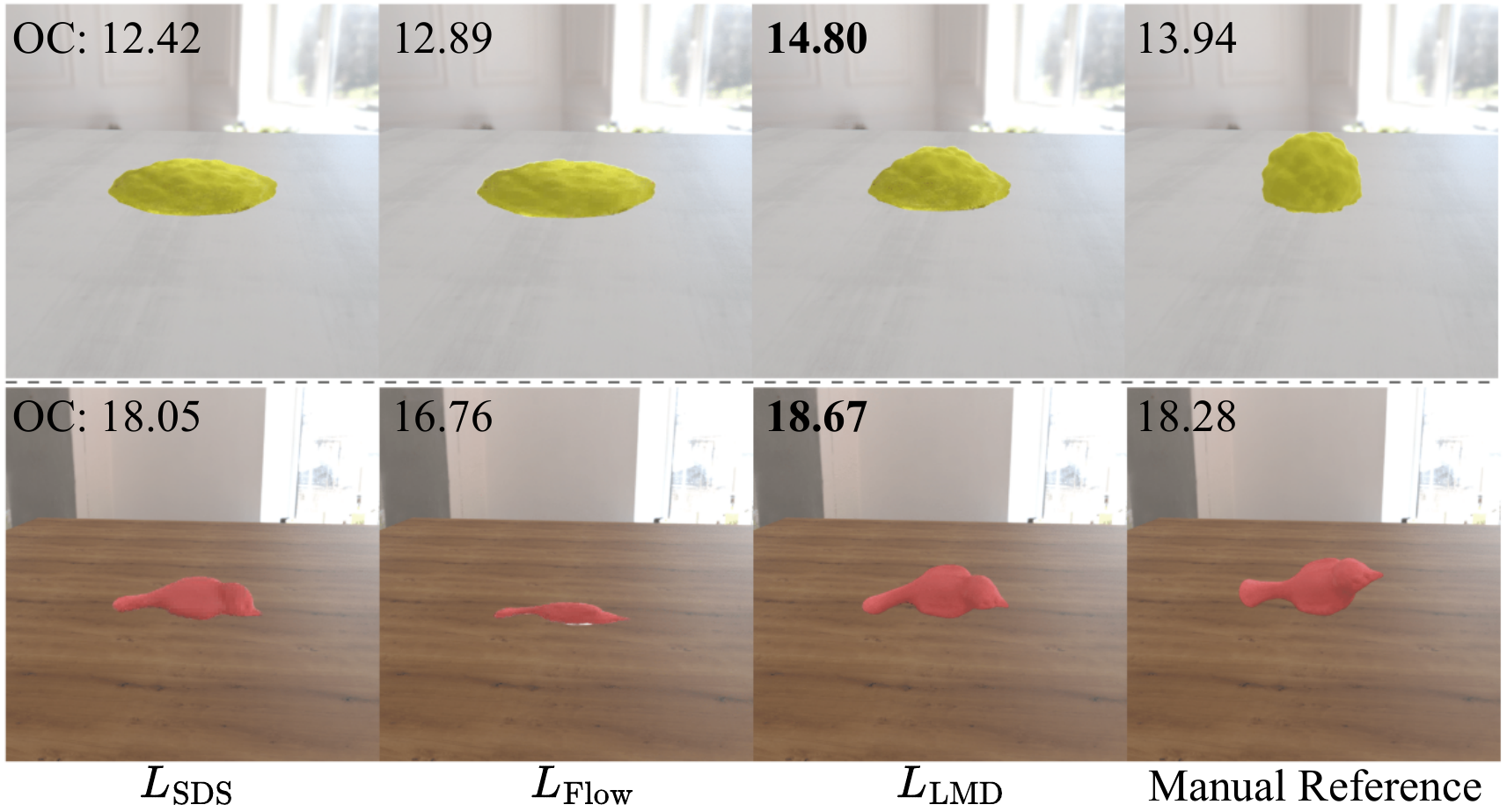}
\caption{\textbf{Ablation of $L_{\mathrm{LMD}}$}. The corresponding Overall Consistency (OC $\times 10^{-2} \uparrow$) scores are also provided.} 
\label{fig:ablation_lmd}
\end{figure}

\noindent\textbf{\emph{Impact of LMD.}}  
To demonstrate the advantage of our learnable motion distillation (LMD) objective, we compare three losses: 1) the optical flow loss \(L_{\mathrm{Flow}}\) from PhysFlow, which extracts flow from generated videos; 2) the SDS loss \(L_{\mathrm{SDS}}\) (Eq.~\ref{eq:sds}); and 3) our LMD loss \(L_{\mathrm{LMD}}\) (Eq.~\ref{eq:lmd}). In Fig.~\ref{fig:ablation_lmd}, because the prompts specify the same materials as in the raw simulation data (Playdoh: Top, plasticine; Bird: Bottom, elastic), we use PAC‑NeRF’s manually tuned outputs~\cite{li2023pac} as a coarse reference for material behavior, even though it is  originally developed for system identification.
 The models optimized with \(L_{\mathrm{LMD}}\) align most closely with this reference, confirming its superior material‑parameter optimization (see quantitative results in Tab.~\ref{tab:ablation-metrics} and additional ablations of \(L_{\mathrm{LMD}}\) in Suppl.).
Note,  the results on PhysFlow's project page~\cite{liu2025physflow_web} closely resemble the manual references since they perform system identification on human-designed objects trained on the paired GT videos. 

\begin{table}[!htb]
  \centering  \resizebox{\linewidth}{!}{
    \begin{tabular}{cccc}
      \toprule
      Ours Init. $\rightarrow$ PhysFlow Init. & Ours Init. w/o Optim. & $L_\mathrm{LMD} \rightarrow L_\mathrm{SDS}$ & \textbf{Ours} \\
      \midrule
      18.04 & 18.05 & 18.13 & \textbf{18.18} \\
      \bottomrule
    \end{tabular}
  } 
  \caption{
    \textbf{Ablation Results.} Reported using OC ($\times 10^{-2}\uparrow$).
  }
\label{tab:ablation-metrics}
\end{table}

\begin{figure}[h]
\centering
\includegraphics[width=0.95\linewidth]{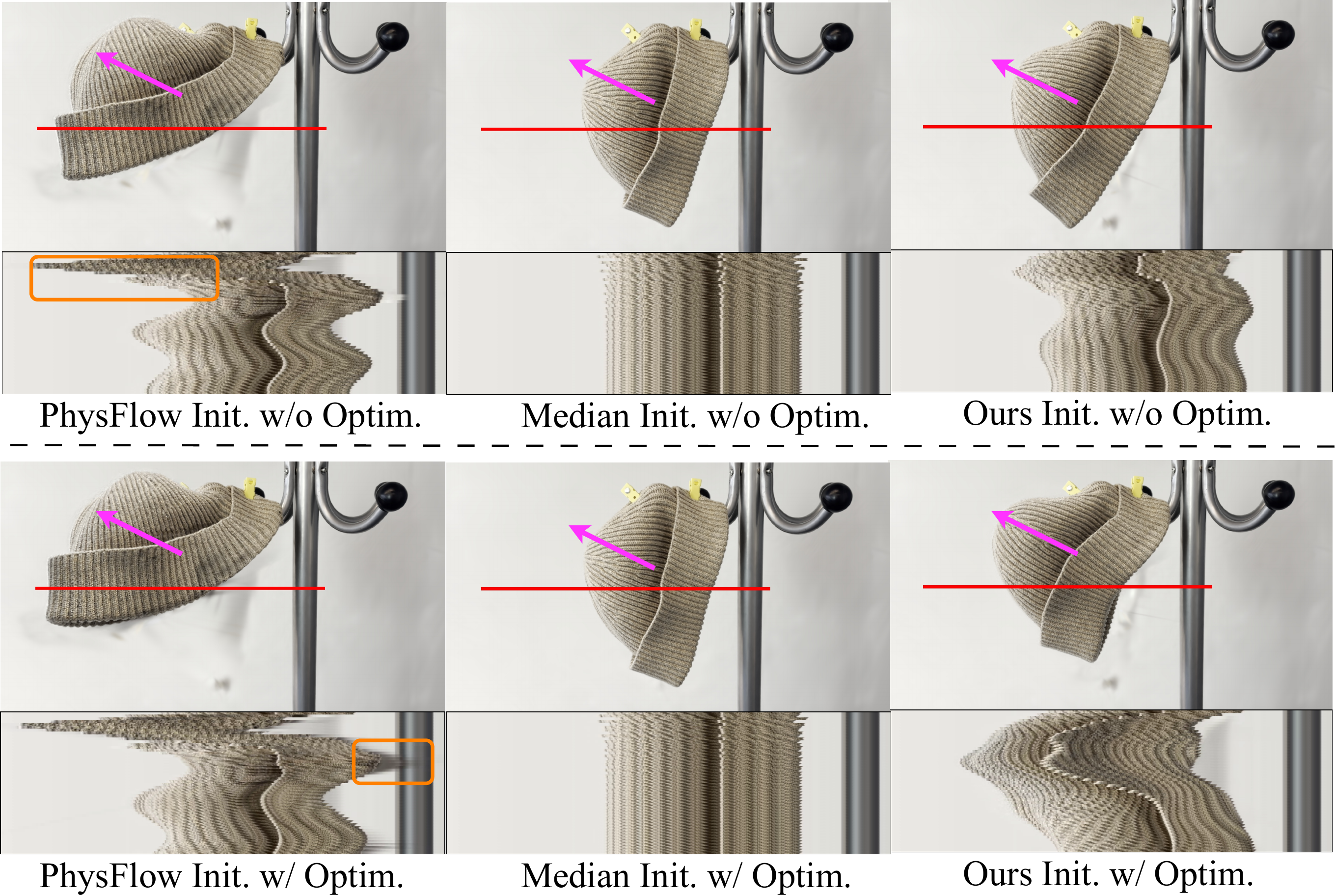}
\caption{\textbf{Initialization Ablation.} We display oscillations with space–time slices, with time on the y-axis and the object’s cross‑section (red lines in the top ``Object'') on the x-axis, revealing both oscillations in amplitude and frequency. 
}
\label{fig:ablation_init}
\end{figure}

\noindent\textbf{\emph{Impact of Initialization.}} Our method employs material-specific range prompts, in contrast to PhysFlow’s naïve LLM reasoning. 
To validate this, in Fig.~\ref{fig:ablation_init} we simulate the Hat scene under six conditions: 
PhysFlow initialization with (using LMD) and without optimization, 
our initialization with (using LMD) and without optimization, 
and median-value initialization with (using LMD) and without optimization (using the median of the constrained value ranges).  
PhysFlow's initialization results in exaggerated early deformations and, even after optimization, still produces unrealistic artifacts, as highlighted in the orange rectangles. Median-value initialization is highly sensitive to the upper bound. 
In this case, a large Young’s modulus of $2\times10^{11}$ leads to overly rigid behavior with minimal deformation failing adhere to the provided force and textual description, even after optimization. In contrast, our approach provides a stable and plausible starting point (see Tab.~\ref{tab:quant-metrics}). When combined with LMD, it yields material dynamics that are both accurate and visually convincing. These results demonstrate that our range prompts guide the LLM to select appropriate parameters values rather than hallucinating spurious guesses.

\begin{figure}[!htb]
\centering
\includegraphics[width=0.95\linewidth]{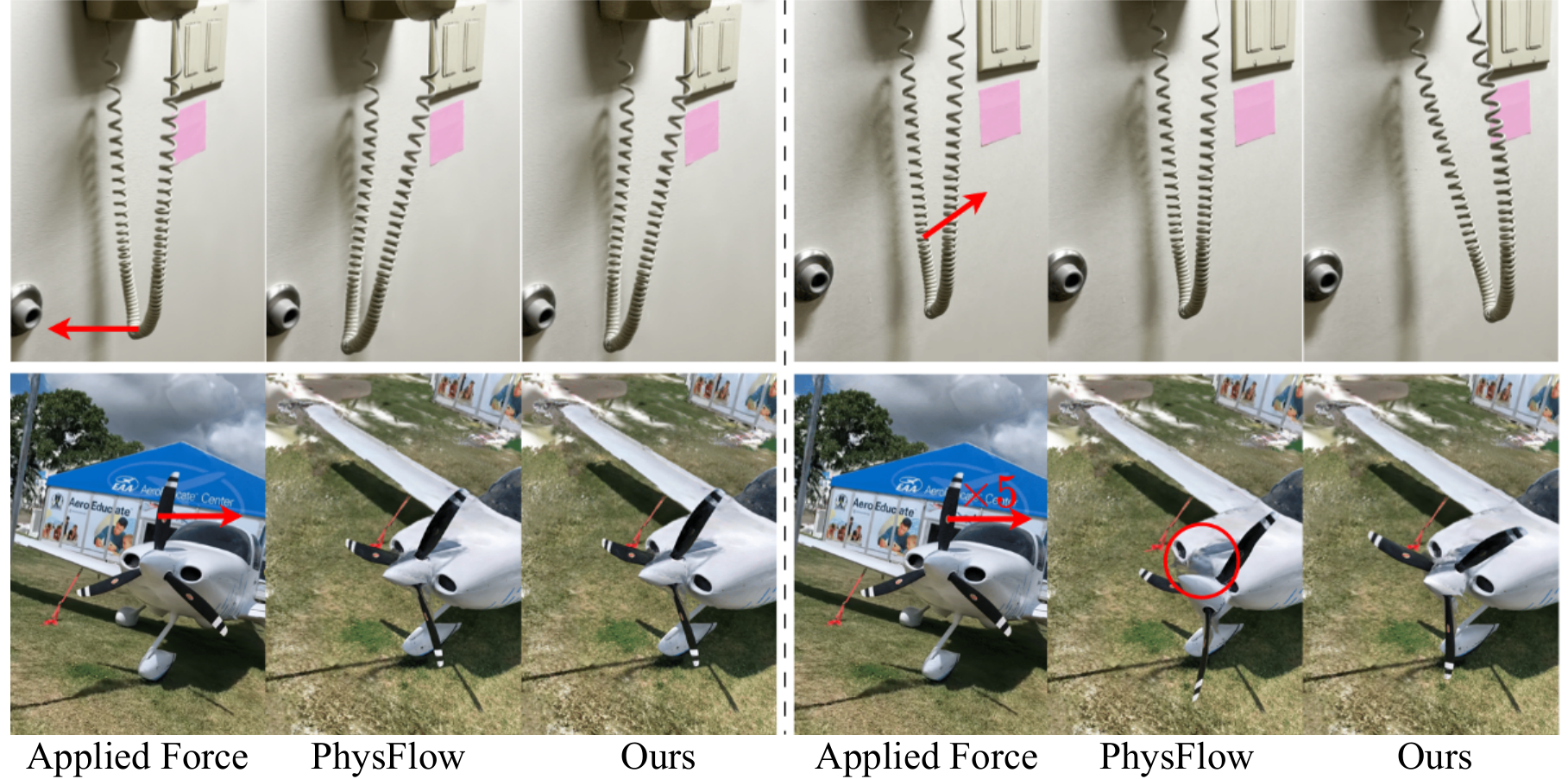}
\caption[\textbf{Robustness to Varying Simulation Conditions}]{%
\textbf{Robustness to Varying Simulation Conditions.}
(Top) Varying force directions and  viewpoint direction.
(Bottom) The propeller rotation speed is increased fivefold.}
\label{fig:robust}
\end{figure}
\vspace{1mm}
\noindent\textbf{\emph{Robustness to Varying Simulation Conditions.}}
We evaluate each scene under varied external forces and slight camera perturbations.
In Fig.~\ref{fig:robust} (Top), the telephone is subjected to a different force direction and a shifted viewpoint. Our method still produces the correct elastic deformation guided by the prompt: \textit{``The telephone cord is gently vibrating”}, whereas PhysFlow remains nearly static as the generated videos themselves fail to provide the correct motion that serves as the GT labels for its loss $(L_{\mathrm{Flow}})$ (see Figure~A6 in Supplement).  In Fig.~\ref{fig:robust} (Bottom),  we quintuple the plane's propeller’s rotational speed. Thanks to the high Young’s modulus and yield strength of metals from our ranges, our simulation preserves structural integrity and yield stress. In contrast, PhysFlow detaches the propeller (red circle) even with the same prompt: \textit{``The plane propeller is spinning”}.

\begin{figure}[!htb]
\centering
\includegraphics[width=0.9\linewidth]{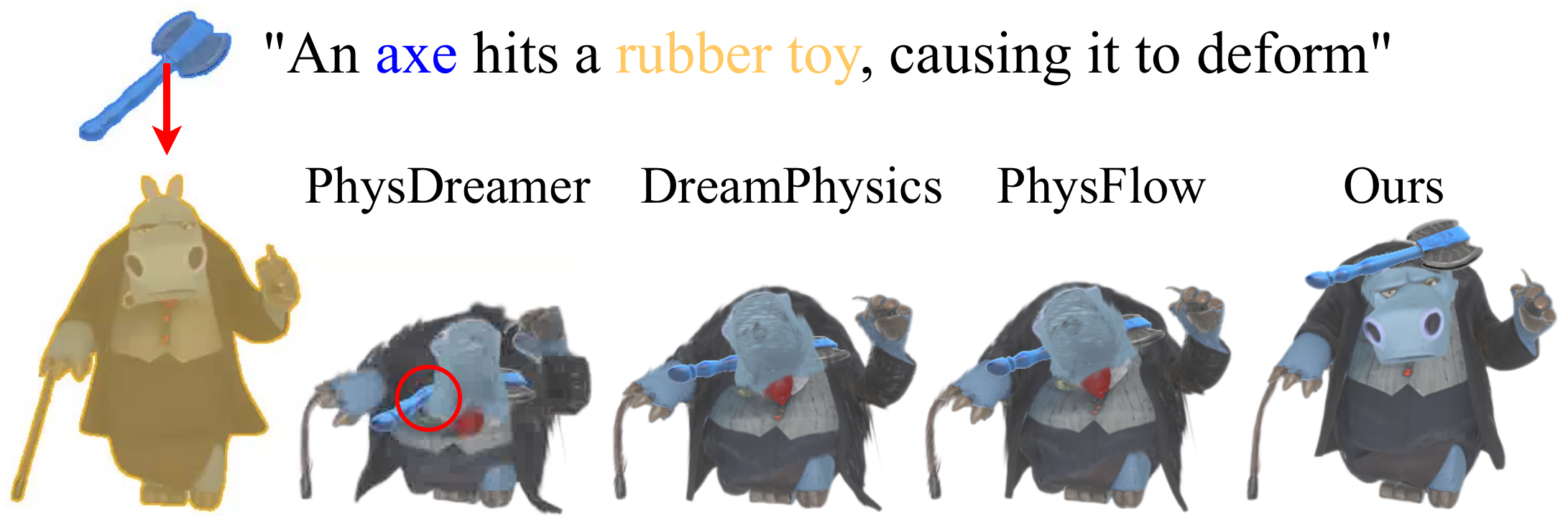}
\caption{
  \textbf{Extension to Heterogeneous Materials.} Our method produces fewer visual artifacts compared to others. 
} 
\label{fig:multi_objects}
\end{figure}

\vspace{1mm}
\noindent\textbf{\emph{Extension to Heterogeneous Materials.}} Our approach extends seamlessly to multi‑object scenes with different materials. By combining GS segmentation \cite{decoupledGaussian} with SAM2 \cite{ravisam}, a multimodal LLM cab infer each object’s material properties from a single text prompt. As shown in Fig.~\ref{fig:multi_objects}, the axe is treated as metal and the toy as elastic rubber.  In our MPM framework, every particle carries its own material parameters, yielding high‑fidelity heterogeneous dynamics. For example, the rubber toy deforms under load while the axe remains rigid. Elastic artifacts in PhysDreamer are highlighted in red (see our attached video). Note, at the time of writing, multi-object simulation code for OmniPhysGS was not available. Adapting our pipeline to complex applications like cinematics or video games involves further engineering beyond this work’s scope.

\section{Conclusion}
\noindent \textbf{MotionPhysics} is a novel framework that uses video diffusion models and multimodal LLMs to drive 3D dynamic scene simulations guided by simple text prompts. A learnable motion distillation module extracts clean motion cues, while an LLM-based embedding initializes material-specific parameter priors, enabling high-fidelity, physically grounded animations. In future, we aim to support fully automatic configuration of fine-grained simulations from text and extend our motion distillation loss to other animation tasks, such as character rigging and deformation.

\vspace{1mm}
\noindent\textbf{Limitations.} Our method does not model shadow effects, which could improve visual realism. Additionally, while our estimated parameters enable plausible simulations, they are not intended for accurate real-world material measurement.

\section{Acknowledgments}
The authors wish to acknowledge the generous support of this research by Huawei Technologies Co. Ltd.
\bibliography{aaai2026}

\clearpage
\frenchspacing  
\setlength{\pdfpagewidth}{8.5in} 
\setlength{\pdfpageheight}{11in} 
%

%

\definecolor{codebg}{rgb}{0.95, 0.95, 0.95}
\definecolor{keywordcolor}{rgb}{0.26, 0.38, 0.67}
\definecolor{commentcolor}{rgb}{0.33, 0.47, 0.09}
\definecolor{stringcolor}{rgb}{0.64, 0.08, 0.08}
\renewcommand\thesection{\Alph{section}}
\setcounter{table}{0}
\renewcommand{\thetable}{A\arabic{table}}
\setcounter{figure}{0}
\renewcommand{\thefigure}{A\arabic{figure}}

\lstset{%
	basicstyle={\footnotesize\ttfamily},
	numbers=left,numberstyle=\footnotesize,xleftmargin=2em,
	aboveskip=0pt,belowskip=0pt,%
	showstringspaces=false,tabsize=2,breaklines=true}
\floatstyle{ruled}
\newfloat{listing}{tb}{lst}{}
\floatname{listing}{Listing}
%
\pdfinfo{
/TemplateVersion (2026.1)
}

\newcommand\norm[1]{\left\lVert#1\right\rVert}

\setcounter{secnumdepth}{0} 
\title{MotionPhysics: Learnable Motion Distillation for Text-Guided Simulation}
\author{
    Written by AAAI Press Staff\textsuperscript{\rm 1}\thanks{With help from the AAAI Publications Committee.}\\
    AAAI Style Contributions by Pater Patel Schneider,
    Sunil Issar,\\
    J. Scott Penberthy,
    George Ferguson,
    Hans Guesgen,
    Francisco Cruz\equalcontrib,
    Marc Pujol-Gonzalez\equalcontrib
}

\setcounter{table}{0}
\renewcommand{\thetable}{A\arabic{table}}
\setcounter{figure}{0}
\renewcommand{\thefigure}{A\arabic{figure}}

\lstdefinestyle{mypython}{
  backgroundcolor=\color{codebg},
  basicstyle=\ttfamily\footnotesize,
  breaklines=true,
  frame=single,
  columns=fullflexible,
  keepspaces=true,
  language=Python,
  keywordstyle=\color{keywordcolor}\bfseries,
  commentstyle=\color{commentcolor}\itshape,
  stringstyle=\color{stringcolor},
  showstringspaces=false,
  tabsize=4
}

\let\titleold\title
\providecommand{\thetitle}{}
\renewcommand{\title}[1]{%
  \titleold{#1}%
  \renewcommand{\thetitle}{#1}%
}

\def\maketitlesupplementary{%
  \newpage
  \twocolumn[%
    \centering
    \Large
    \textbf{\thetitle}\\
    \vspace{0.5em}Supplementary Material\\
    \vspace{1.0em}
  ]
}
\title{MotionPhysics: Learnable Motion Distillation for Text-Guided Simulation}

\maketitlesupplementary

\section{Organization}
This supplementary material includes an overview of the constitutive models, our complete LLM-based material initialization prompt with  parameter ranges for different material types, details of our motion-invariant GS perturbation, boundary conditions, qualitative comparisons to other methods, additional ablations, and additional user study results, including voting statistics and  screenshots of the questionnaire. We recommend watching our supplementary video for the full dynamic simulation results. 
\section{Constitutive Models}
Here, we briefly describe the physical models used in our simulation setup. The reader is encouraged to refer to the relevant sources for additional  information. 

\subsection{Governing Equations in Continuum Mechanics}

Continuum mechanics describes material motion using a time-dependent deformation map $\mathbf{x} = \psi(\mathbf{X}, t)$, which maps each material point $\mathbf{X}$ in the undeformed reference configuration $\Omega^0$ to its position $\mathbf{x}$ in the deformed configuration $\Omega^t$ at time $t$. The local motion and shape changes are characterized by the \emph{deformation gradient}:
\begin{equation}
  \mathbf{F}(\mathbf{X}, t) = \nabla_\mathbf{X} \psi(\mathbf{X}, t),
\end{equation}
which encodes local stretching, rotation, and shearing effects~\cite{bonet1997nonlinear}.

The evolution of the deformation field $\psi$ is governed by conservation of mass and conservation of momentum. The latter yields the elastodynamics equation:
\begin{equation}
  \rho_0 \ddot{\psi} = \nabla \cdot \mathbf{P} + \rho_0 \mathbf{b},
  \label{eq:elastodynamics}
\end{equation}
where:
\begin{itemize}
  \item $\ddot \psi$ denote the acceleration of a material point.
  \item $\rho_0$ is the reference (initial) mass density.
  \item $\mathbf{P}$ is the first Piola-Kirchhoff stress tensor.
  \item $\mathbf{b}$ is the body force per unit mass.
\end{itemize}

This PDE system does not describe the system completely. 
Constitutive laws are required to relate the stress $\mathbf{P}$ to the material state, i.e., to how the material is deformed and, in the case of inelasticity, to its deformation history.

\subsection{Constitutive Laws: Elasticity and Plasticity}

Constitutive laws describe how materials respond to deformation. In general, for solids, following prior works \cite{jiang2016material,ma2023learning}, we distinguish the following types:
\begin{itemize}
    \item \textbf{Elastic law:} Relates the (possibly elastic) deformation gradient $\mathbf{F}^e$ to the stress, governing the reversible (recoverable) part of the deformation.
    \item \textbf{Plastic flow and integration (return mapping):} When the stress state exceeds a material-dependent yield criterion, permanent (irreversible) deformation occurs. Yield conditions, flow rules, and hardening laws specify the evolution of the plastic state. In practice, a \emph{return mapping} algorithm is used at each time step to enforce these constraints numerically.
\end{itemize}

The generic forms are:
\begin{align}
    &\text{Elastic Law:} && \mathbf{P} = \widehat{P}(\mathbf{F}^e), \\
    &\text{Yield Criterion:} && f_Y(\mathbf{P}) < 0,\\
    &\text{Plastic Update:} && \mathbf{F}^e \leftarrow \mathrm{ReturnMap}(\mathbf{F}^e).
\end{align}

\subsection{Classical Material Models and Return Mapping}

Table~\ref{tab:constitutive-laws} summarizes classical material models and the return mapping (plastic integration) schemes commonly implemented in computational mechanics and in this work. 
Below we provide  a concise but precise overview of each constitutive model and its plastic integration scheme used:

\begin{table}[t]
\resizebox{\linewidth}{!}{
\begin{tabular}{l l l}
\toprule
\textbf{Material Type} & \textbf{Elastic Law} &  \textbf{Return Mapping (Plastic Integration)} \\
\midrule
Elastic & Fixed Corotated Elasticity & Identity mapping \\
Plasticine & StVK Elasticity & von Mises Plasticity with Damage \\
Metal & StVK Elasticity & von Mises Plasticity \\
Foam & StVK Elasticity & Viscoplastic Return Mapping\\
Sand & StVK Elasticity & Drucker-Prager Plasticity \\
Newtonian fluid & Compressible Neo-Hookean Elasticity & Identity mapping \\
Non‑Newtonian fluid & StVK Elasticity & Herschel-Bulkley Plasticity \\
\bottomrule
\end{tabular}
}
\caption{Representative constitutive models and corresponding plastic integration schemes (return mapping), following the implementation in PhysFlow~\cite{liu2025physflow}. The elastic law governs the recoverable stress response, while the return mapping algorithm projects the trial state back onto the plastic yield surface.}
\label{tab:constitutive-laws}
\end{table}

\paragraph{Fixed Corotated Elasticity}
\begin{align}
    &\mathbf{P} = 2\mu(\mathbf{F}-\mathbf{R})+\lambda J (J-1)\mathbf{F}^{-T},\\[3pt]
    &\text{No plastic yields}, \quad f_{Y}(\mathbf{P})\equiv -1< 0,\\[3pt]
    &\mathbf{F}^e \leftarrow \mathbf{F}, \quad \text{Identity mapping (no evolution)}.
\end{align}

\paragraph{St.Venant–Kirchhoff Elasticity (StVK)}
\begin{align}
    &\boldsymbol{\tau} = \mathbf{U}\left[2\mu\,\boldsymbol{\epsilon}+\lambda\,\text{tr}(\boldsymbol{\epsilon})\mathbf{I}\right]\mathbf{U}^{T},\quad \boldsymbol{\epsilon}=\log(\mathbf{\Sigma}),\\
    &f_Y(\tau) \text{ depends on specific plastic model below},\\
    &\mathbf{F}^{e}\leftarrow\text{ReturnMap}(\mathbf{F}^{e}).
\end{align}

\paragraph{von Mises Plasticity}
(von Mises yield criterion and radial return mapping)
\begin{align}
    &f_Y(\tau) = \|\mathbf{s}\| - \sqrt{\frac{2}{3}}\sigma_Y<0,\quad \mathbf{s}=\text{dev}(\boldsymbol{\tau}),\\[3pt]
    &\mathbf{F}^{e}\leftarrow \text{RadialReturnMap}_\text{vonMises}(\mathbf{F}^{e}).
\end{align}

\paragraph{von Mises Plasticity with Damage}
(Yield stress decreases with accumulated plastic strain)
\begin{align}
    &f_Y(\tau) = \|\mathbf{s}\| - \sqrt{\frac{2}{3}}\sigma_Y(\epsilon^p)<0,\quad \frac{d\sigma_Y}{d\epsilon^p}<0,\\[3pt]
    &\mathbf{F}^{e},\sigma_Y\leftarrow \text{RadialReturnMap}_\text{DamagedVM}(\mathbf{F}^{e},\epsilon^p).
\end{align}

\paragraph{Viscoplastic Return Mapping (Foam)}
(Viscous regularized von Mises yield law)
\begin{align}
    &f_Y(\tau) = \|\mathbf{s}\| - \sqrt{\frac{2}{3}}\sigma_Y<0,\\
    &s = s^{\text{trial}} - \frac{y}{1+\frac{\eta}{2\mu\Delta t}}\frac{s^{\text{trial}}}{\|s^{\text{trial}}\|},\\[3pt]
    &\mathbf{F}^{e}\leftarrow \text{ViscoplasticReturnMap}(\mathbf{F}^{e},\eta).
\end{align}

\paragraph{Drucker-Prager Plasticity (Sand)}
(Frictional cone yield criterion)
\begin{align}
    &f_Y(\tau)=\|\mathbf{s}\|+\alpha\,\text{tr}(\tau)-k<0,\\[3pt]
    &\mathbf{F}^{e}\leftarrow\text{ReturnMap}_\text{DruckerPrager}(\mathbf{F}^{e},\alpha,k).
\end{align}

\paragraph{Compressible Neo-Hookean Elasticity (Newtonian Fluid)}
(Weakly compressible elasticity; no distinct plastic step)
\begin{align}
    &\mathbf{P}=\frac{\mu}{J^{2/3}}(\mathbf{F}-J^{-1/3}\mathbf{R}) +\frac{\kappa}{2}(J^2 -1)\mathbf{F}^{-T},\\[3pt]
    &\text{No plastic yields}, \quad f_{Y}(\mathbf{P})\equiv -1< 0,\\[3pt]
    &\mathbf{F}^{e}\ \leftarrow\mathbf{F},\quad\text{Identity mapping}.
\end{align}

\paragraph{Herschel-Bulkley Plasticity (Non-Newtonian Fluid)}
(Rate-dependent viscoplastic fluid)
\begin{align}
f_Y(\tau)&=\|\mathbf{s}\|-\sqrt{\frac{2}{3}}\sigma_Y<0,\\[3pt]
s &= s^{\text{trial}}- \frac{s^{\text{trial}}-\sqrt{\frac{2}{3}}\sigma_Y}{1+\frac{\eta}{2\mu\Delta t}}\frac{s^{\text{trial}}}{\|s^{\text{trial}}\|},\\[3pt]
\mathbf{F}^{e} &\leftarrow \text{ReturnMap}_\text{HerschelBulkley}(\mathbf{F}^{e},\eta,\sigma_Y).
\end{align}

This comprehensive set of constitutive models allows our model to capture rich elastoplastic behaviors for various real-world material types, supporting high-fidelity simulations.

\section{LLM-based Material Initialization}

Following
PhysFlow \cite{liu2025physflow}, we consider seven representative material types,
each characterized by typical parameter sets to cover a broad
range of commonly encountered materials:
\begin{enumerate}
   \item \textbf{Elastic:} Young’s modulus $E$ and Poisson’s ratio $\nu$
   \item \textbf{Plasticine:} $E$, $\nu$,and yield stress $\tau_Y$
   \item \textbf{Metal:} $E$, $\nu$, and yield stress $\tau_Y$
   \item \textbf{Foam:} $E$, $\nu$, and plastic viscosity $\eta$
   \item \textbf{Sand:} friction angle $\theta_{\mathrm{fric}}$
   \item \textbf{Newtonian fluid:} fluid viscosity $\mu$ and bulk modulus $\kappa$
   \item \textbf{Non-Newtonian fluid:} shear modulus $\mu$, bulk modulus $\kappa$, yield stress $\tau_Y$, and plastic viscosity $\eta$
\end{enumerate}

We consulted physical parameter range limits from standard material property handbooks~\cite{callister_materials,ashby_materials,mitchell_soil,gibson_ashby_foam,crc_handbook} and the chosen ranges are shown in Table~\ref{tab:mpm_material_params}.
\begin{table*}[ht]
  \centering
\adjustbox{width=0.6\linewidth}{%
    \begingroup
    \renewcommand{\arraystretch}{1.2}
    \setlength{\tabcolsep}{8pt}
    \begin{tabular}{l c c l}
      \toprule
      \textbf{Material Type} & \textbf{Parameter} & \textbf{SI Unit} & \textbf{Value Range} \\
      \midrule
      Elastic 
        & $E$ (Young's modulus)       & Pa            & $10^7$\,–\,$4\times10^{11}$~\cite{callister_materials} \\
        & $\nu$ (Poisson's ratio)     & (unitless)    & 0.1\,–\,0.5~\cite{callister_materials} \\
      \midrule
      Plasticine 
        & $E$                         & Pa            & $1\times10^6$\,–\,$5\times10^6$~\cite{ashby_materials} \\
        & $\nu$                       & (unitless)    & 0.3\,–\,0.4~\cite{ashby_materials} \\
        & $\tau_Y$ (yield stress)     & Pa            & $5\times10^3$\,–\,$2\times10^4$~\cite{ashby_materials} \\
      \midrule
      Metal 
        & $E$                         & Pa            & $4.5\times10^{10}$\,–\,$4.0\times10^{11}$~\cite{callister_materials} \\
        & $\nu$                       & (unitless)    & 0.25\,–\,0.35~\cite{callister_materials} \\
        & $\tau_Y$                    & Pa            & $1\times10^7$\,–\,$2\times10^9$~\cite{callister_materials} \\
      \midrule
      Foam 
        & $E$                         & Pa            & $10^3$\,–\,$10^7$~\cite{gibson_ashby_foam} \\
        & $\nu$                       & (unitless)    & 0\,–\,0.3~\cite{gibson_ashby_foam} \\
        & $\tau_Y$                    & Pa            & $1\times10^4$\,–\,$1\times10^6$~\cite{gibson_ashby_foam} \\
        & $\eta$                      & (unitless)    & 0.1\,–\,1~\cite{gibson_ashby_foam} \\
      \midrule
      Sand 
        & $\theta_{\mathrm{fric}}$    & °             & 27\,–\,45°~\cite{mitchell_soil} \\
      \midrule
      Newtonian fluid 
        & $\mu$ (viscosity)           & Pa·s          & $10^{-3}$\,–\,$10$~\cite{crc_handbook} \\
        & $\kappa$ (bulk modulus)     & Pa            & $1\times10^9$\,–\,$5\times10^9$~\cite{crc_handbook} \\
      \midrule
      Non-Newtonian 
        & $\mu$                       & Pa·s          & $10^{-3}$\,–\,$10^3$~\cite{crc_handbook} \\
        & $\kappa$                   & Pa            & $1\times10^9$\,–\,$5\times10^9$~\cite{crc_handbook} \\
        & $\tau_Y$                   & Pa            & $1$\,–\,$2\times10^3$~\cite{ashby_materials} \\
        & $\eta$                     & (unitless)    & 0.1\,–\,1~\cite{ashby_materials} \\
      \midrule
      Common 
        & $\rho$                      & kg·m$^{-3}$   & 10\,–\,$2.3\times10^4$~\cite{crc_handbook,wikipedia_mat} \\
      \bottomrule
    \end{tabular}
    \endgroup
  }
\caption{Material parameters, their SI units, and value ranges for the seven simulated material types (with sources).}
\label{tab:mpm_material_params}  
\end{table*}

We utilize OpenAI’s GPT-4 \cite{achiam2023gpt}, a multimodal large language model (LLM), to infer material types and physical parameters from a textual simulation prompt with an optional image reference. Below is our complete inquiry prompt used in our pipeline:
\begin{quote}
\noindent \textbf{Inputs:}
\begin{itemize}
    \item \textbf{Textual simulation prompt:} \\
    \texttt{"An axe is hitting the ground."}
    \item \textbf{Reference Image:}
\begin{figure}[htb!]
    \centering
    \includegraphics[width=0.5\linewidth]{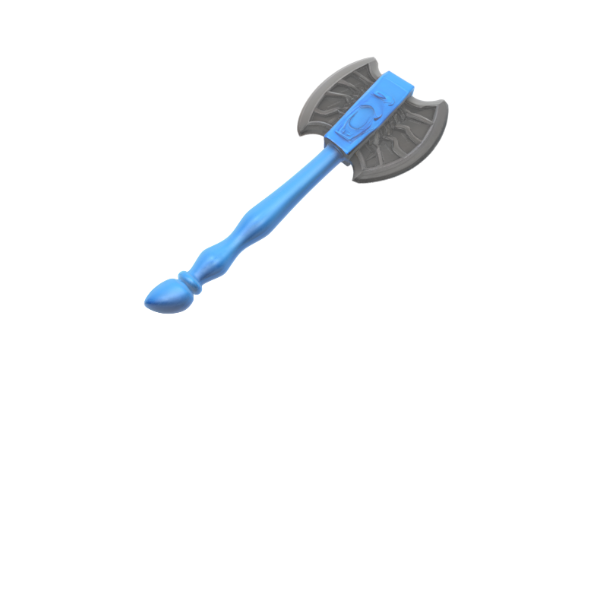}
    \label{fig:enter-label}
\end{figure}
\end{itemize}
\vspace{-20mm}

\textbf{Q:} What is this object? Based primarily on the \textit{textual simulation prompt}, determine the most appropriate material type. Then estimate its density (kg/m³) and relevant physical parameters. Use the \textit{image as secondary visual reference only}.

\noindent\textbf{Warning:} Material types and required parameters (with SI units):
\begin{itemize}
  \item Elastic: $E$ (Pa), $\nu$ (unitless)
  \item Plasticine: $E$, $\nu$, $\tau_Y$ (Pa)
  \item Metal: $E$, $\nu$, $\tau_Y$ (Pa)
  \item Foam: $E$, $\nu$, $\tau_Y$, $\eta$ (unitless)
  \item Sand: $\theta_{fric}$ (°)
  \item Newtonian fluid: $\mu$ (Pa·s), $\kappa$ (kPa)
  \item Non-Newtonian fluid: $\mu$, $\kappa$, $\tau_Y$ (Pa), $\eta$ (unitless)
\end{itemize}

\noindent\textbf{Warning:} All inferred physical parameters \textit{must strictly fall within the following valid ranges}:

\begin{itemize}
  \item \textit{Elastic:}
    \begin{itemize}
      \item $E$: $1 \times 10^7$ – $4 \times 10^{11}$ Pa
      \item $\nu$: 0.1 – 0.5
    \end{itemize}
  \item \textit{Plasticine:}
    \begin{itemize}
      \item $E$: $1 \times 10^6$ – $5 \times 10^6$ Pa
      \item $\nu$: 0.3 – 0.4
      \item $\tau_Y$: $5 \times 10^3$ – $2 \times 10^4$ Pa
    \end{itemize}
  \item \textit{Metal:}
    \begin{itemize}
      \item $E$: $4.5 \times 10^{10}$ – $4.0 \times 10^{11}$ Pa
      \item $\nu$: 0.25 – 0.35
      \item $\tau_Y$: $1 \times 10^7$ – $2 \times 10^9$ Pa
    \end{itemize}
  \item \textit{Foam:}
    \begin{itemize}
      \item $E$: $1 \times 10^3$ – $1 \times 10^7$ Pa
      \item $\nu$: 0 – 0.3
      \item $\tau_Y$: $1 \times 10^4$ – $1 \times 10^6$ Pa
      \item $\eta$: 0.1 – 1
    \end{itemize}
  \item \textit{Sand:} $\theta_{fric}$: 27 – 45°
  \item \textit{Newtonian fluid:}
    \begin{itemize}
      \item $\mu$: $1 \times 10^{-3}$ – 10 Pa·s
      \item $\kappa$: $1 \times 10^6$ – $5 \times 10^6$ kPa
    \end{itemize}
  \item \textit{Non-Newtonian fluid:}
    \begin{itemize}
      \item $\mu$: $1$ – $1 \times 10^6$ mPa·s
      \item $\kappa$: $1 \times 10^6$ – $5 \times 10^6$ kPa
      \item $\tau_Y$: 1 – $2 \times 10^3$ Pa
      \item $\eta$: 0.1 – 1
    \end{itemize}
  \item \textit{Density} $\rho$: 10 – $2.3 \times 10^4$ kg/m³
\end{itemize}

\noindent\textbf{Respond in this exact format:}
\vspace{1mm}

\begin{lstlisting}[style=mypython]
{
  "material_type": "...",
  "density": ...,
  "E": ...,
  "nu": ...,
  "tau_Y": ...,
  "mu": ...,
  "kappa": ...,
  "eta": ...,
  "theta_fric": ...
}
\end{lstlisting}

\noindent\textit{Only include fields relevant to the inferred material type.}

\noindent \textbf{LLM Answer:}
\vspace{1mm}
\begin{lstlisting}[style=mypython]
{
  "material_type": "Metal",
  "density": 7850,
  "E": 2.1e11,
  "nu": 0.30,
  "tau_Y": 2.5e8
}
\end{lstlisting}
\end{quote}

Moreover, following PhysFlow, we apply the same linear or logarithmic scaling to GPT-4’s predictions, followed by the same clamping to ensure numerical stability in our differentiable MLS-MPM simulations.

Our proposed approach allows large language models (LLMs) to directly infer simulation parameters from natural language prompts and visual cues, providing a plausible initialization for subsequent parameter optimization aimed at reproducing the scenario described by the user. 
\section{Motion-Invariant GS Perturbation}

To promote motion distillation during training, we add controlled noise to both the Gaussian centers $\mathbf{x}_g$ and their spherical harmonics coefficients $\mathcal{S}_g$:

\begin{equation}
\tilde{\mathbf{x}}_g = \mathbf{x}_g + \epsilon\,\mathcal{N}(\mathbf{0}, \mathbf{I}), 
\end{equation} 
\begin{equation}
\tilde{\mathcal{S}}_g = \mathcal{S}_g + 2\epsilon\,\mathcal{N}(\mathbf{0}, \mathbf{I}),
\end{equation} 
where $\epsilon=0.1$ sets the perturbation scale. These noisy Gaussians are used only for computing the motion-distillation loss during optimization and are discarded during simulation and evaluation. By mildly varying geometry and appearance, this targeted augmentation drives the model to learn motion features that remain stable under small GS perturbations, resulting in more robust and semantically meaningful motion representations.

\section{Boundary Conditions and Force Application}

We adopt a diverse set of boundary condition modules and force application schemes to support complex material behavior and scene interaction. Our framework includes six types of boundary conditions and five distinct external force modules as shown in Table~\ref{tab:boundary_conditions}, allowing precise spatiotemporal control over particle and grid behavior.
We conclude our simulation settings for real-world dataset cases shown in Table~\ref{tab: forces_conditions}, which differ from those provided in the supplementary materials of PhysFlow.

\begin{table*}[htb!]
\centering
\adjustbox{width=0.8\linewidth}{
\begin{tabular}{lll}
\toprule
\textbf{Category} & \textbf{Mode / Name} & \textbf{Description and Parameters} \\
\midrule

\multirow{2}{*}{\textbf{Grid-based}} 
    & \texttt{clamp\_grid} (Box) & Cuboidal velocity Dirichlet clamp. \textbf{Parameters:} \texttt{v}, \texttt{bounds}, \texttt{axis}. \\
    & \texttt{bounding\_box} & Global bounding box constraint. Projects velocity inward; optional friction term. \\

\midrule

\multirow{4}{*}{\textbf{Surface Collider}} 
    & \texttt{surface\_collider} – \texttt{sticky} & Sticks on contact: sets $\mathbf{v} = 0$. Friction must be zero. \\
    & \texttt{surface\_collider} – \texttt{slip} & Zero normal velocity; tangential velocity retained. \\
    & \texttt{surface\_collider} – \texttt{frictional} (default) & One-way velocity projection using Coulomb friction (inward only). \\
    & \texttt{surface\_collider} – \texttt{cut} & Dampens or clamps velocity in a spatial region; removes material. \\

\specialrule{1.2pt}{2pt}{2pt} 

\multirow{5}{*}{\textbf{Force}} 
    & \texttt{add\_constant\_force} & Applies constant body force within a region. Time-dependent activation. \\
    & \texttt{add\_impulse} & Applies instantaneous impulse to selected particles. \\
    & \texttt{force\_particles\_translation} & Enforces translation on particles. Selective by axis and group. \\
    & \texttt{force\_particles\_rotation} & Enforces rotation on particles. Optional group selection. \\
    & \texttt{release\_particles} & Sequentially releases frozen particles by index, based on time. \\

\bottomrule
\end{tabular}
}
\caption{Overview of boundary condition and external force modules}
\label{tab:boundary_conditions}
\end{table*}

\begin{table*}[htb]
\centering
\adjustbox{width=0.7\linewidth}{
\begin{tabular}{lll}
\toprule
\textbf{Scene} & \textbf{Force Module (Direction)} & \textbf{Duration} \\
\midrule
Alocasia   & \texttt{add\_impulse} $(0.44, 0, 0)$                             & $2$ substeps \\
Carnation  & \texttt{add\_impulse} $(-0.1, 0, 0)$                             & $1$ substep \\
Hat        & \texttt{add\_impulse} $(2, -2, 2)$                               & $1$ second \\
Telephone  & \texttt{add\_constant\_force} $(15, 15, -15)$                    & $0.75$ seconds \\
Fox        & \texttt{add\_impulse} $(0, -0.5, 0.25) \rightarrow (0, 0, -0.5) \rightarrow (0, 0.5, 0.25)$ & $1 \rightarrow 1 \rightarrow 1$ substeps \\
Plane      & \texttt{force\_particles\_rotation} $-50 \rightarrow -5$         & $0.8 \rightarrow 1$ seconds \\
Kitchen    & \texttt{force\_particles\_translation} $(0, 0, 0.1)$             & $5$ seconds \\
Jam        & \texttt{force\_particles\_translation} $(0, 0.3, 0) \rightarrow (0.3, 0, 0)$ & $2.7 \rightarrow 2.3$ seconds \\
Sandcastle & \texttt{release\_particles} $n_{\text{layer}} = 200$            & $5$ seconds \\
\bottomrule
\end{tabular}
}
\caption{Forces and corresponding durations applied in simulations on the real-world dataset. Durations are reported in substeps or seconds.}
\label{tab: forces_conditions}
\end{table*}

\section{Additional Qualitative Results}
\begin{figure*}[!htb]
\centering
\includegraphics[width=\textwidth]{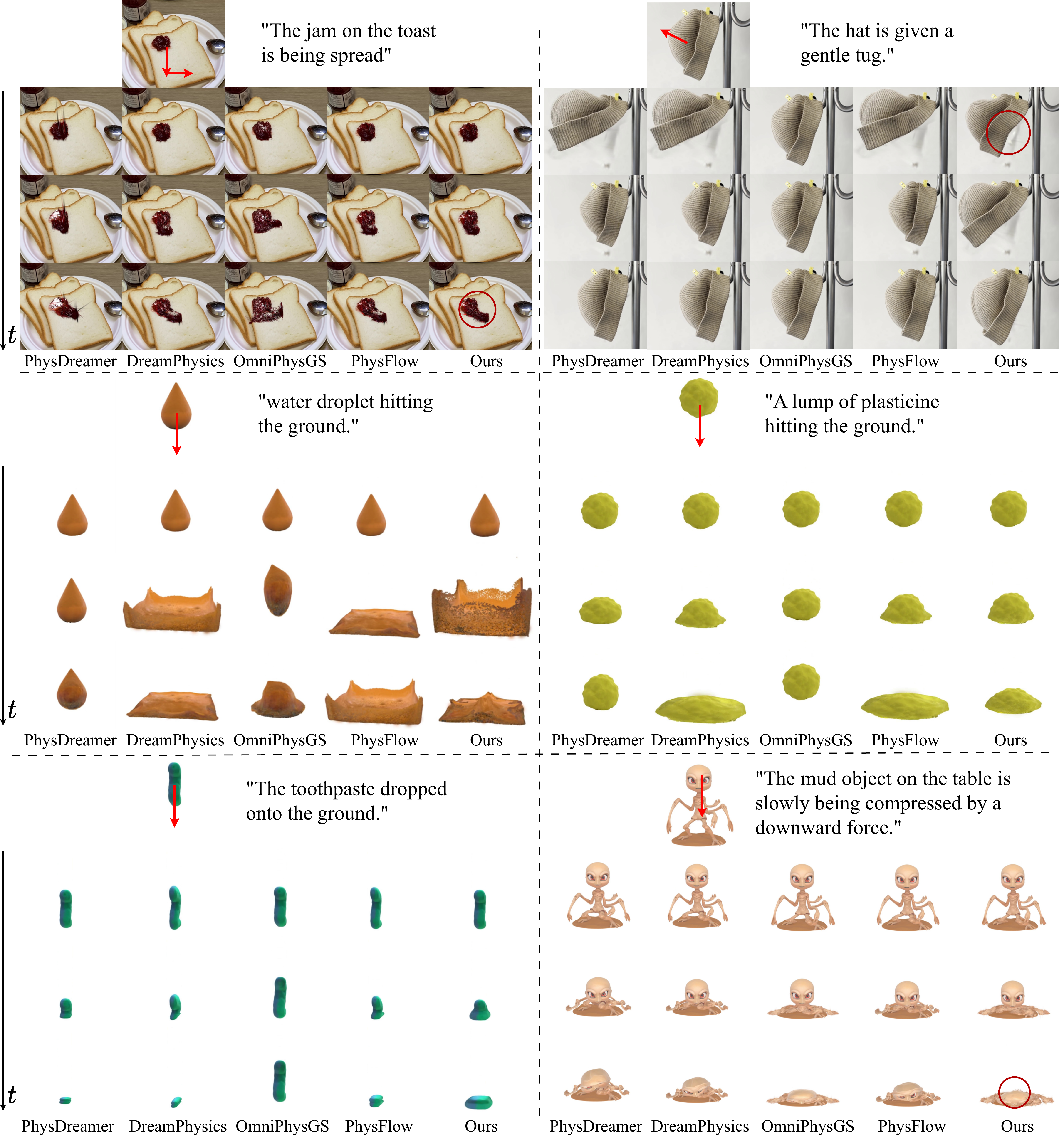}
\caption{\textbf{Qualitative Evaluation.} We compare all baselines including PhysDreamer \cite{zhang2024physdreamer}, DreamPhysics \cite{huang2025dreamphysics}, OmniPhysGS \cite{lin2025omniphysgs}, and PhysFlow \cite{liu2025physflow}, on diverse simulation cases, including real-world scenes (Jam, Hat), synthetic human-like objects (Droplet, Playdoh, Toothpaste), and generated objects (Alien). Red arrows indicate the input force, and red circles highlight noticeable regions. Our method achieves the most faithful dynamic simulations, closely matching both the texture prompt and the applied force.}
\vspace{5mm}
\label{fig:qualitative}
\end{figure*}

\begin{figure*}[!htb]
\centering
\includegraphics[width=\linewidth]{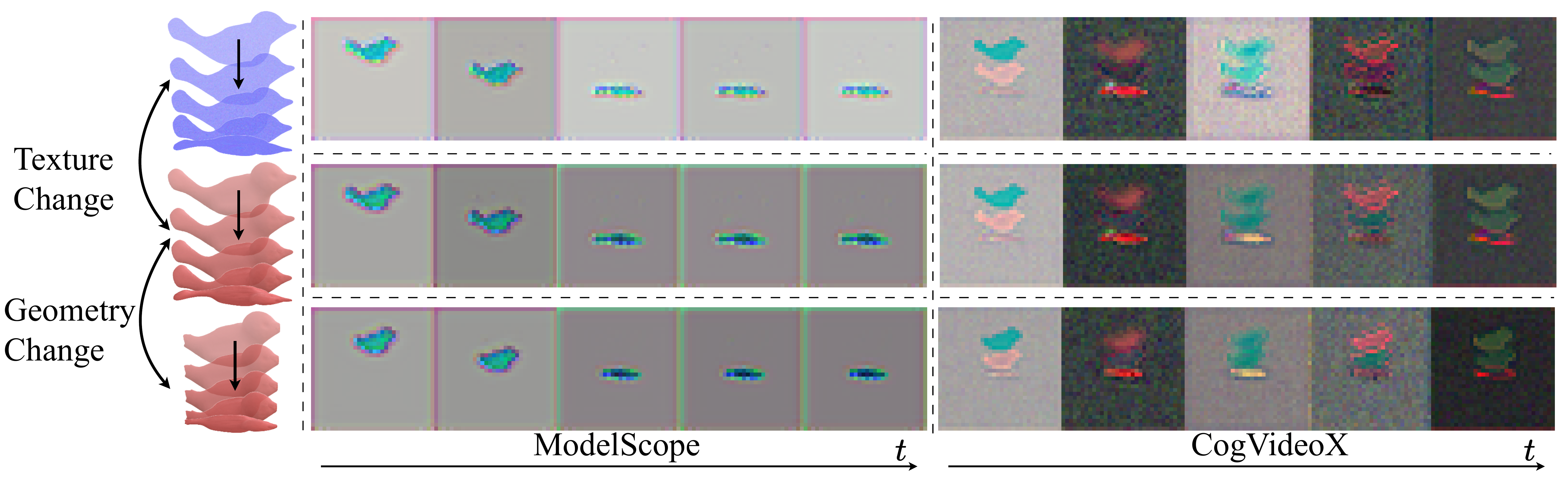}
\caption{\textbf{Structure Similarity Supplment.} Video latents visualized via PCA preserve the global structure across different video encoders in video diffusion models, specifically ModelScope (Left) and CogVideoX (Right), for both texture (Top) and geometry (Bottom) variants, which differ only in local, part‑level details. These variants follow the same motion sequences simulated using identical initialization parameters estimated by our LLM before subsequent optimizations.}
\label{fig:latent_visualizatio_another}
\end{figure*}

We provide additional qualitative comparisons in Figure~\ref{fig:qualitative}, along with further ablation studies presented below.
\paragraph{\emph{Motion Pattern Consistency}.}
Variations in texture or shape, when subjected to identical simulated motions under the same initial physical parameters, yield globally consistent latent codes generated by the video encoder. As shown in Figure~\ref{fig:latent_visualizatio_another}, this consistency in motion patterns emerges across different video encoders used in text-to-video generation models, including both ModelScope \cite{wang2023modelscope} (Left) and CogVideoX \cite{yang2024cogvideox} (Right).

\begin{figure*}[!htb]
\centering
\includegraphics[width=\linewidth]{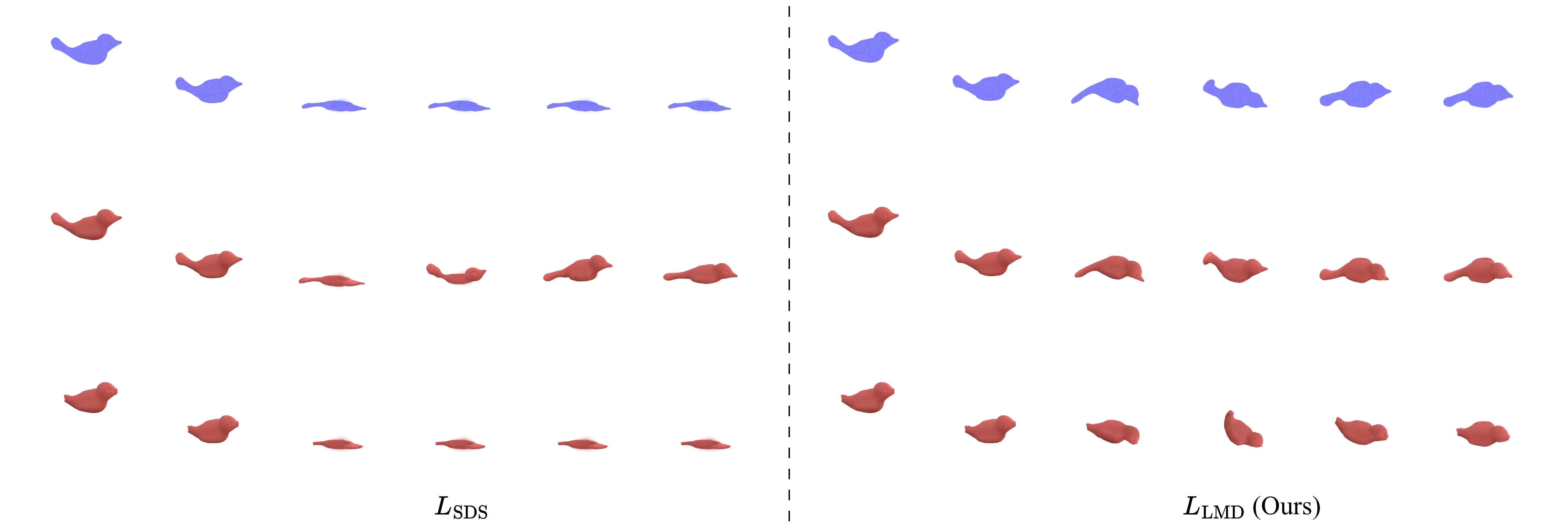}
\caption{\textbf{Additional Ablation of $L_\mathrm{LMD}$.}
We compare our learnable motion distillation loss $L_\mathrm{LMD}$ with the conventional score distillation loss $L_\mathrm{SDS}$, using identical initialization parameters and the prompt \textit{“An elastic bird is hitting the ground.”} After optimization, $L_\mathrm{LMD}$ achieves consistent elastic motion patterns across both texture and geometry variants, faithfully reflecting the text prompt. In contrast, $L_\mathrm{SDS}$ fails to produce the expected dynamics, yielding divergent behaviors for the two variants (see our supplementary video for the full dynamics).
}
\label{fig:Additional_Ablation_LMD}
\end{figure*}

\paragraph{\emph{Additional Ablation of $L_\mathrm{LMD}$}.}
We present the simulated sequence corresponding to Fig.~3 (Right) in the main paper. As shown in Figure~\ref{fig:Additional_Ablation_LMD}, our learnable motion distillation loss $L_\mathrm{LMD}$ consistently distills motion patterns across both geometry and texture variations, whereas optimization with $L_\mathrm{SDS}$ produces divergent simulations that fail to adhere to the user-provided inputs.

\begin{figure}[!htb]
\centering
\includegraphics[width=\linewidth]{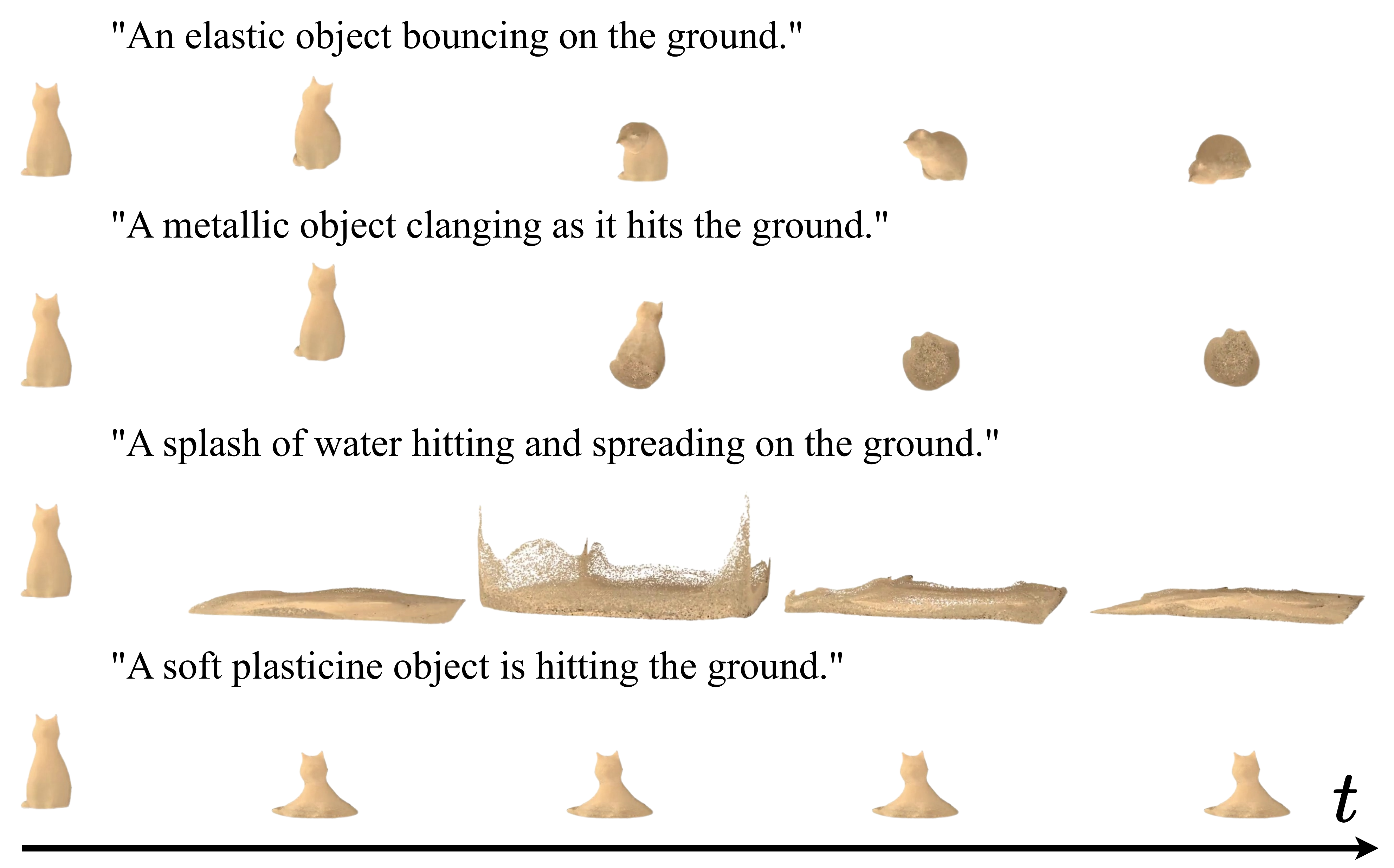}
\caption{\textbf{Robustness to Varying Text Prompts.} Our method successfully simulates a diverse range of dynamic effects guided by different text prompts, using the same synthetic Cat scene. From top to bottom, the results successfully demonstrate behaviors corresponding to elastic, metal, Newtonian fluid, and plasticine materials.}
\label{fig:robust:text_prompts}
\end{figure}

\paragraph{\emph{Robustness to Varying Text Prompts}.} 
We evaluate how faithfully each method can follow different simulation textual descriptions for the same object. As illustrated in Figure~\ref{fig:robust:text_prompts}, our approach reliably produces the intended effects across diverse prompts, confirming that it minimizes biases from geometry or texture and instead prioritizes the motion specified in the user-provided text prompt.

\begin{figure}[!htb]
\centering
\includegraphics[width=\linewidth]{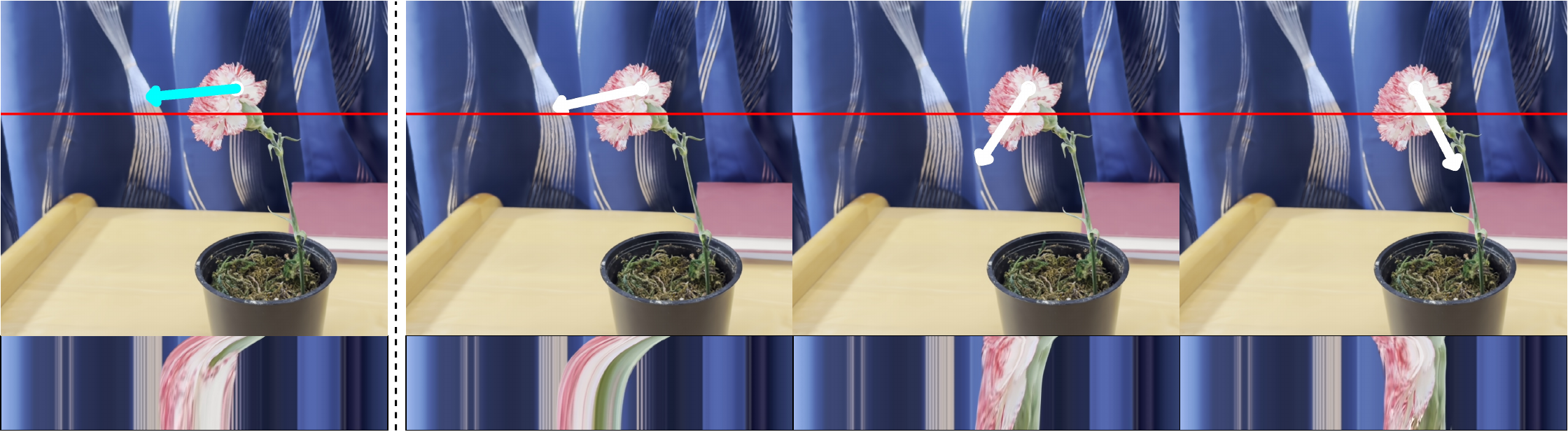}
\caption{%
\textbf{Interactive Evaluation.} We visualize the resulting oscillations via space–time slices shown on the bottom.
}
\label{fig:evaluations}
\end{figure}

\paragraph{\emph{Interactive Evaluation}.} Following prior work \cite{zhang2024physdreamer}, we train our model only once on a single object. During evaluation, we introduce new simulation conditions to generate diverse interaction scenarios (Figure~\ref{fig:evaluations}). After the initial training (cyan arrow, Left), the object is able to deform under novel applied forces (white arrows, Right).

\begin{figure}[!htb]
\centering
\includegraphics[width=\linewidth]{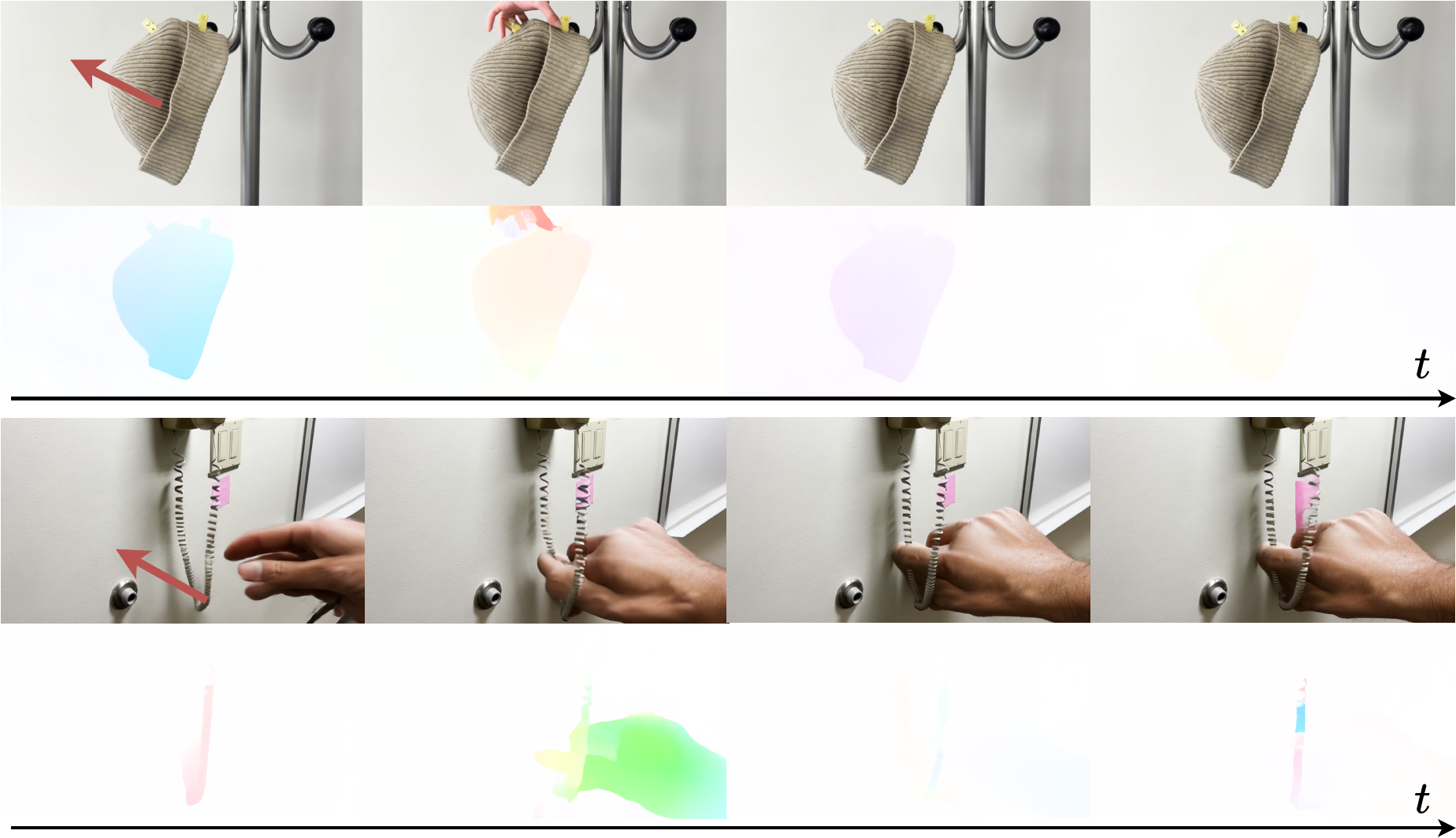}
\caption{\textbf{PhysFlow Failure.} Visualization of the generated video frames and their corresponding optical flow for the Hat (Top) and Telephone (Bottom) scenes. The optical flow does not reflect the true physical motion required by the simulation.
}
\label{fig:physflow_failure}
\end{figure}

\paragraph{\emph{Failure Analysis of PhysFlow}.} We further analyze the failure cases of PhysFlow. Although PhysFlow is also based on the intuition of extracting motion signals, its method relies on optical flow estimated from generated videos. However, these generated videos often fail to reflect the actual motion dictated by simulation conditions (e.g., applied forces), even in real-world scenarios, as illustrated in Figure~\ref{fig:physflow_failure}.

\section{User Study Details}
\begin{table*}[ht]
\centering
\resizebox{\linewidth}{!}{
\setlength{\tabcolsep}{4pt}
\renewcommand{\arraystretch}{0.95}
\begin{tabular}{@{}l*{15}{c}|c@{}}
\toprule
\textbf{Physical realism} & Alien & Alocasia & Urchin & Bird & Cat & Cream & Droplet & Hat & Jam & Lego & Plane & Playdoh & Sandcastle & Telephone & Toothpaste & Sum \\
\midrule
Ours over PhysDream    & \textbf{26 / 28} & \textbf{13 / 13} & \textbf{18 / 19} & \textbf{13 / 13} & \textbf{13 / 13} & \textbf{18 / 19} & \textbf{19 / 19} & \textbf{25 / 28} & \textbf{28 / 28} & \textbf{25 / 28} & \textbf{16 / 19} & \textbf{17 / 19} & \textbf{11 / 13} & 3 / 13  & \textbf{27 / 28} & \textbf{272 / 300} \\
Ours over DreamPhysics & \textbf{19 / 19} & \textbf{11 / 19} & \textbf{25 / 28} & \textbf{18 / 28} & \textbf{11 / 19} & \textbf{17 / 19} & \textbf{24 / 28} & \textbf{16 / 19} & \textbf{19 / 19} & \textbf{19 / 19} & \textbf{24 / 28} & \textbf{13 / 13} & 3 / 19  & \textbf{21 / 28} & \textbf{16 / 19} & \textbf{256 / 324} \\
Ours over OmniPhysGS   & \textbf{9 / 13}  & \textbf{18 / 19} & \textbf{19 / 19} & \textbf{19 / 19} & \textbf{19 / 19} & \textbf{13 / 13} & \textbf{19 / 19} & \textbf{16 / 19} & \textbf{12 / 13} & \textbf{13 / 19} & \textbf{19 / 19} & \textbf{24 / 28} & \textbf{26 / 28} & \textbf{19 / 19} & \textbf{13 / 13} & \textbf{258 / 279} \\
Ours over PhysFlow     & \textbf{13 / 19} & \textbf{21 / 28} & \textbf{13 / 13} & \textbf{19 / 19} & \textbf{20 / 28} & \textbf{27 / 28} & \textbf{11 / 13} & \textbf{10 / 13} & \textbf{18 / 19} & \textbf{12 / 13} & \textbf{12 / 13} & \textbf{18 / 19} & 8 / 19  & \textbf{13 / 19} & \textbf{18 / 19} & \textbf{233 / 282} \\
\midrule
\textbf{Prompt adherence} & Alien & Alocasia & Urchin & Bird & Cat & Cream & Droplet & Hat & Jam & Lego & Plane & Playdoh & Sandcastle & Telephone & Toothpaste & Sum \\
\midrule
Ours over PhysDreamer    & \textbf{26 / 28} & \textbf{13 / 13} & \textbf{19 / 19} & \textbf{13 / 13} & \textbf{13 / 13} & \textbf{18 / 19} & \textbf{17 / 19} & \textbf{25 / 28} & \textbf{27 / 28} & \textbf{26 / 28} & \textbf{15 / 19} & \textbf{18 / 19} & \textbf{12 / 13} & 6 / 13  & \textbf{27 / 28} & \textbf{275 / 300} \\
Ours over DreamPhysics & \textbf{18 / 19} & \textbf{14 / 19} & \textbf{26 / 28} & \textbf{23 / 28} & 8 / 19  & \textbf{18 / 19} & \textbf{24 / 28} & \textbf{18 / 19} & \textbf{16 / 19} & \textbf{17 / 19} & \textbf{23 / 28} & \textbf{13 / 13} & 5 / 19  & \textbf{24 / 28} & \textbf{17 / 19} & \textbf{264 / 324} \\
Ours over OmniPhysGS   & \textbf{8 / 13}  & \textbf{18 / 19} & \textbf{19 / 19} & \textbf{19 / 19} & \textbf{19 / 19} & \textbf{13 / 13} & \textbf{19 / 19} & \textbf{15 / 19} & \textbf{11 / 13} & \textbf{14 / 19} & \textbf{19 / 19} & \textbf{26 / 28} & \textbf{25 / 28} & \textbf{19 / 19} & \textbf{13 / 13} & \textbf{257 / 279} \\
Ours over PhysFlow     & \textbf{15 / 19} & \textbf{22 / 28} & \textbf{12 / 13} & \textbf{18 / 19} & \textbf{21 / 28} & \textbf{27 / 28} & \textbf{12 / 13} & \textbf{10 / 13} & \textbf{17 / 19} & \textbf{12 / 13} & \textbf{11 / 13} & \textbf{18 / 19} & \textbf{11 / 19} & \textbf{13 / 19} & \textbf{18 / 19} & \textbf{237 / 282} \\
\bottomrule
\end{tabular}
}
\caption{Explicit vote counts in the user study (our votes / total votes). \textbf{Bold} indicates cases where our method received the majority of the votes. 
}
\label{tab:user-study_votes}
\end{table*}

In our user study, the left/right ordering of videos in each pair was randomized, and all videos were anonymized to prevent participants from knowing which video was ours or one of the baselines based on the ordering. Participants were also unaware of the number of baseline methods involved in the study. 

We evaluated 15 different scenes, each compared against 4 baseline methods, resulting in 60 video pairs. These were randomly divided into four distinct questionnaires, each consisting of 15 pairs, one per scene. Upon accessing the survey via the main link, participants were randomly assigned to one of the four questionnaires.
A total of 79 users participated in the study, resulting in approximately 20 user evaluations per method per scene on average. The raw voting statistics for all test cases are presented in Table~\ref{tab:user-study_votes}.
\begin{figure}[!htb]
  \centering
  \includegraphics[width=\linewidth]{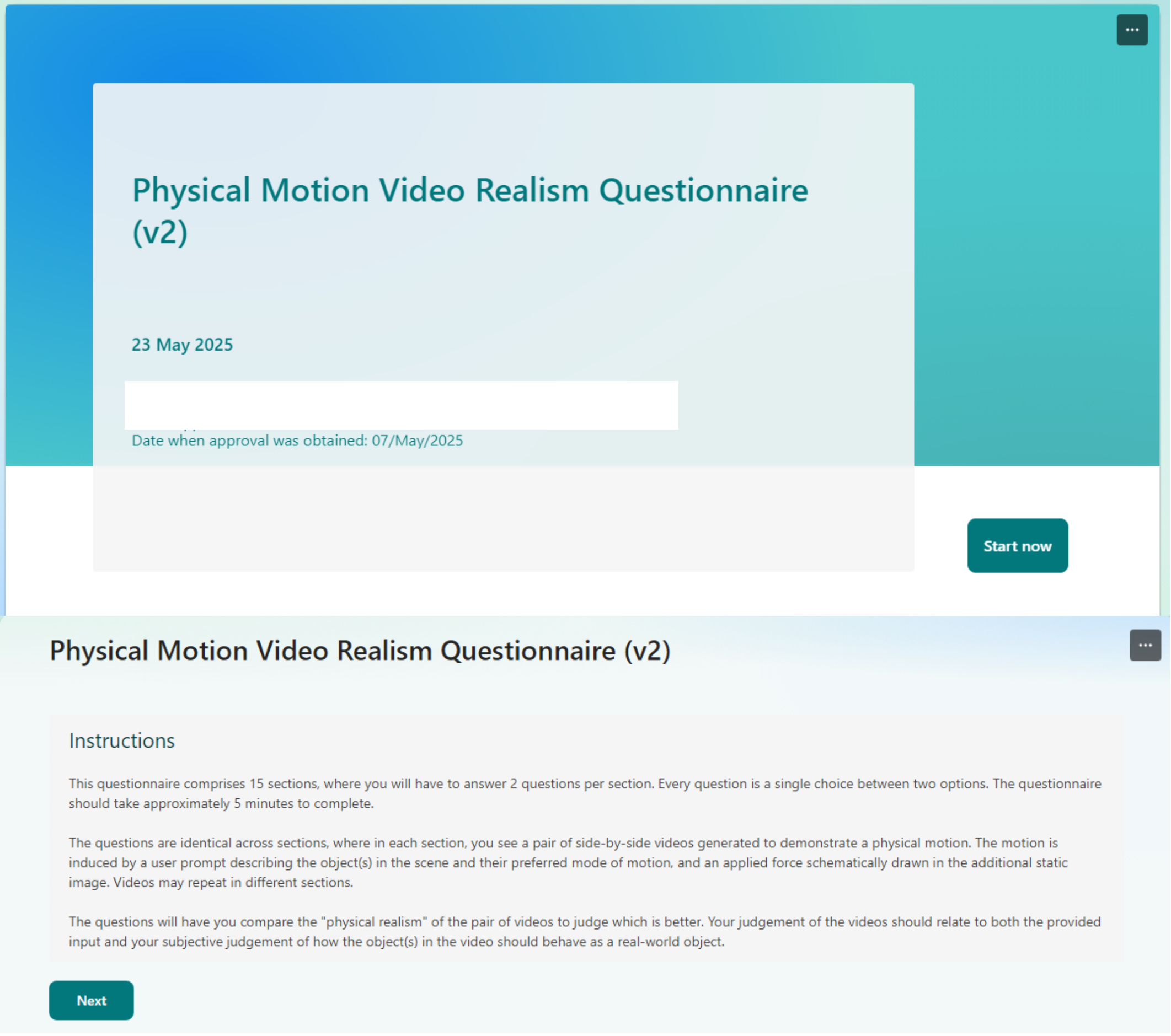}
  \caption{\textbf{Instructions Page.} Screenshot of the user study questionnaire introduction. Personal or institutional information approved by the corresponding research ethics committee has been masked with white rectangles.}
  \label{fig:screenshot_instruction}
\end{figure}

\begin{figure}[!htb]
  \centering
  \includegraphics[width=\linewidth]{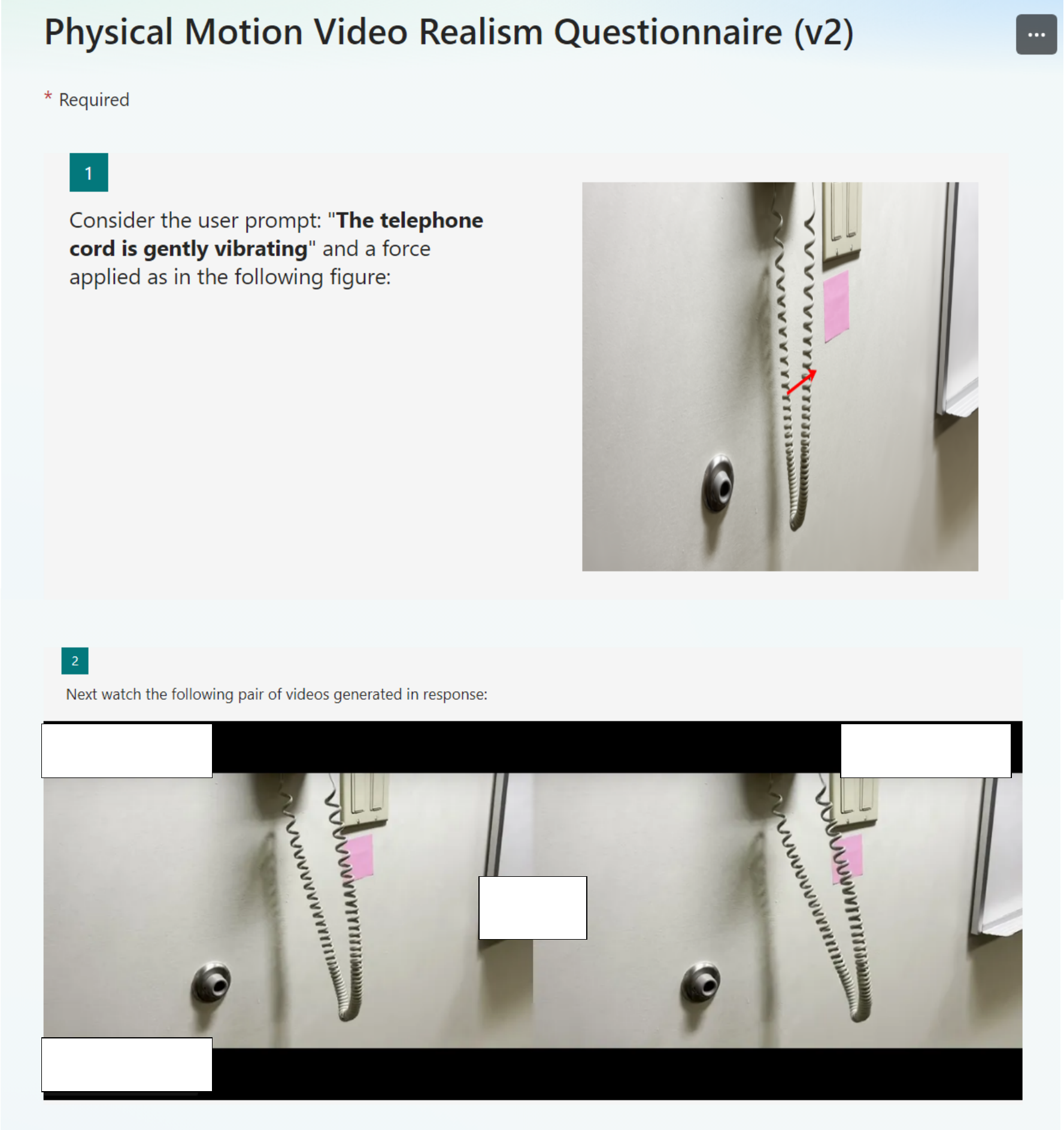}
  \caption{\textbf{Video comparison interface.} Screenshot of the interface displaying the textual prompt, the direction of the applied force, and the paired simulation videos for comparison. Sensitive or identifying information is masked.}
  \label{fig:screenshot_presentation}
\end{figure}

\begin{figure}[!htb]
  \centering
\includegraphics[width=\linewidth]{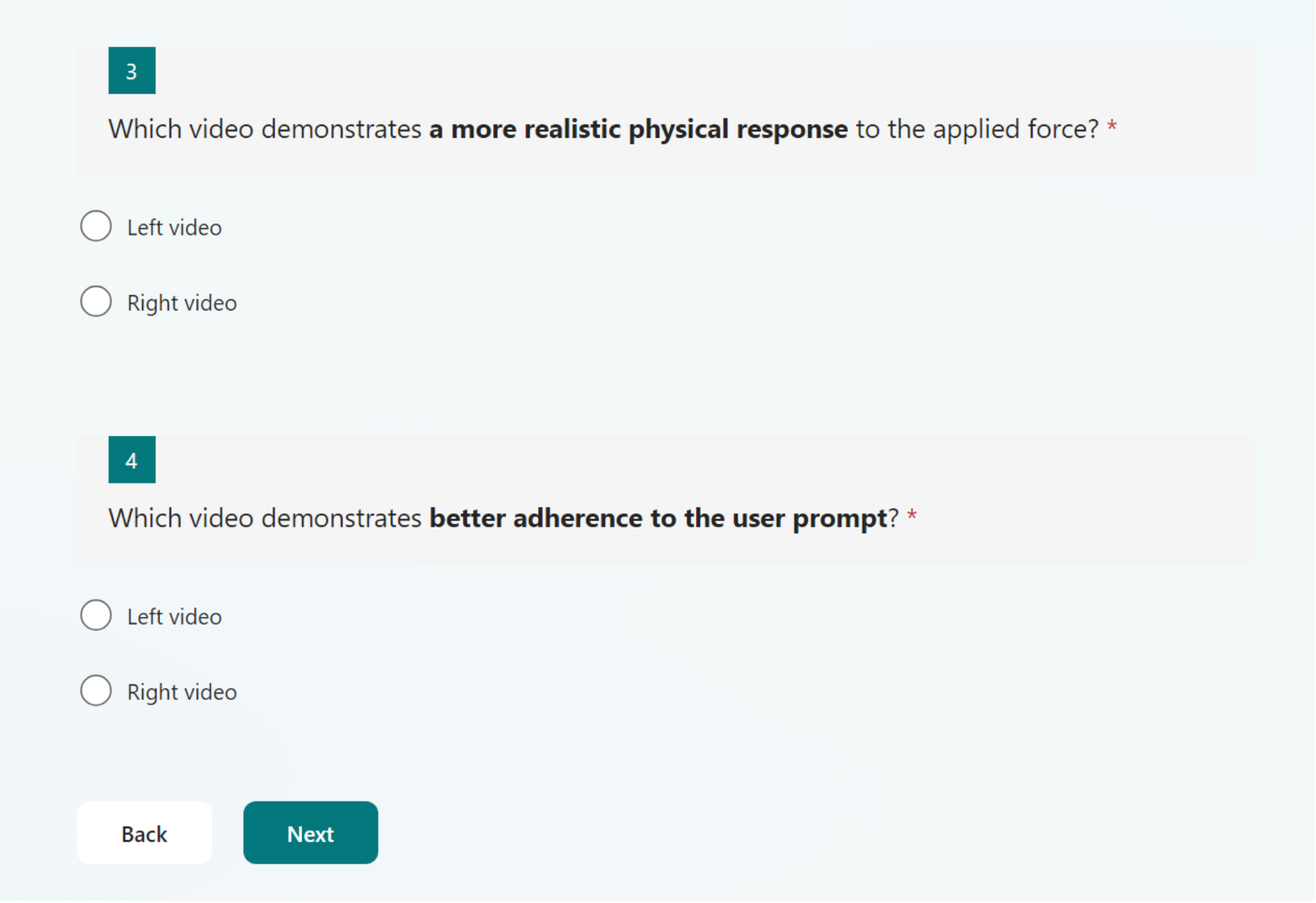}
  \caption{\textbf{Evaluation questions.} Screenshot of the two questions used to collect user preferences, forming the basis of the reported subjective evaluation metrics.}
  \label{fig:screenshot_question}
\end{figure}

We present screenshots from our questionnaire interface in the figures that follow. Figure~\ref{fig:screenshot_instruction} shows the initial instruction page, which includes approved ethics statements, study guidance, and the estimated completion time. Figure~\ref{fig:screenshot_presentation} illustrates the textual prompt, force visualization, and video comparison interface. Figure~\ref{fig:screenshot_question} displays the two evaluation questions asked to collect subjective ratings. 
We attach the raw answers and video assignments to the left/right slots in the respective forms in the submission files (see the \texttt{user\_study} folder for details.).

\begin{figure}[htb!]
  \centering
  \includegraphics[width=\linewidth]{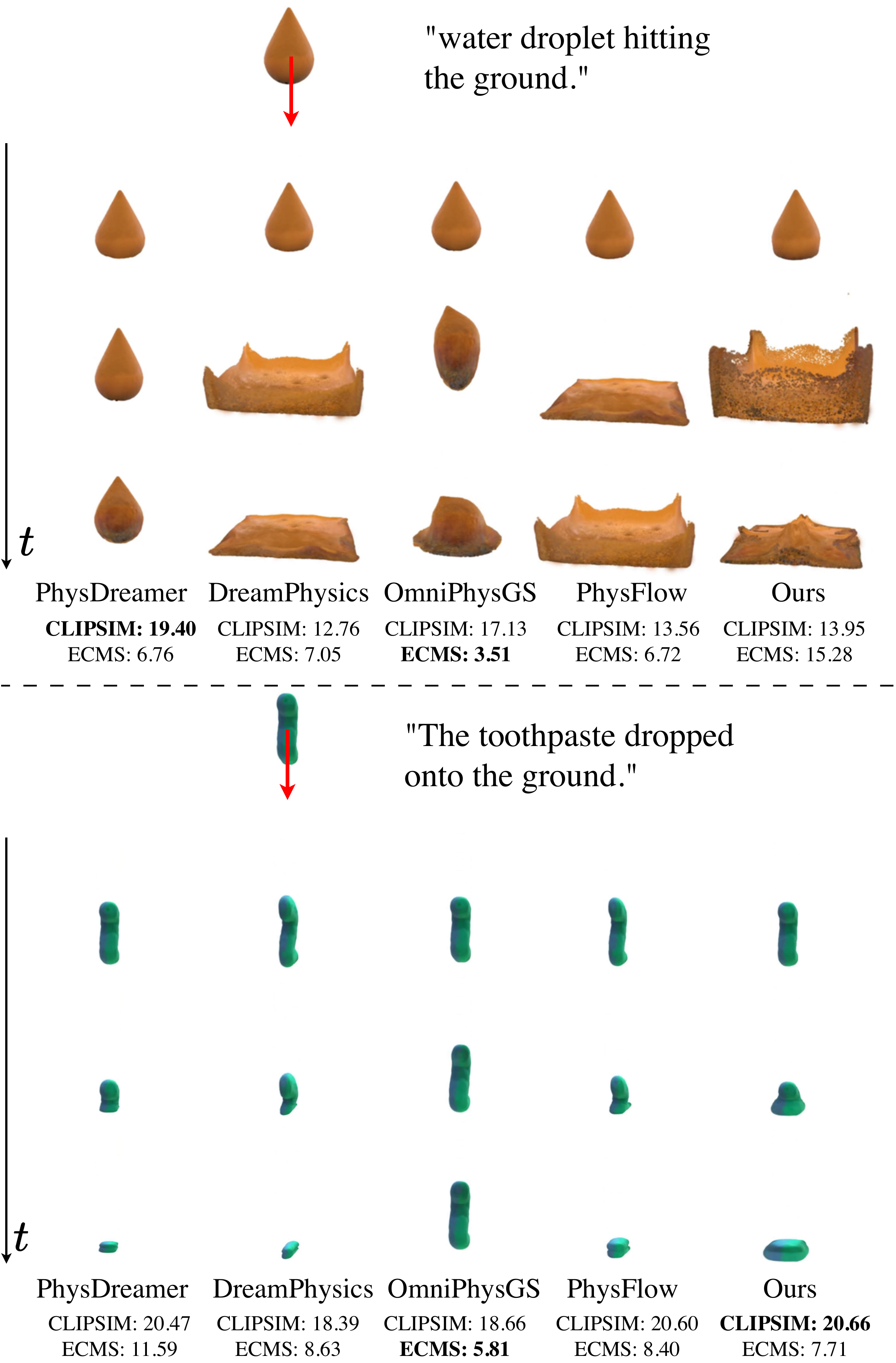}
  \caption{Qualitative results on the Droplet (Top) and Toothpaste (Bottom) scenes with annotated CLIPSIM(\%)($\uparrow$) and ECMS ($\downarrow$) metrics. Best results for each scene are in \textbf{bold}.}
  \label{fig:metric-examples}
\end{figure}

\section{Quantitative Metrics Details}
Here, we provide a detailed analysis of the three metrics, Overall Consistency (OC), CLIPSIM and ECMS, presented in the main paper.

Consider a generated video $I = \{I_1, I_2, \ldots, I_L\}$ for a user prompt $\mathcal{P}_{\text{text}}$, where $I_l$ denotes the frame at time step $l$. To evaluate how well the video adheres to the prompt, we adopt two metrics: Overall Consistency (OC) from VBench~\cite{huang2024vbench} and CLIPSIM~\cite{wu2021godiva}, defined as follows:

\begin{equation}
\textrm{OC}(I, \mathcal{P}_{\text{text}}) = \frac{\left\langle \textrm{ViCLIP}(I), \textrm{ViCLIP}(\mathcal{P}_{\text{text}}) \right\rangle}{\left| \textrm{ViCLIP}(I) \right| \left| \textrm{ViCLIP}(\mathcal{P}_{\text{text}}) \right|},
\end{equation}

where \textrm{ViCLIP}~\cite{wang2023internvid} denotes the joint embedding model for text and video.

\begin{equation}
\textrm{CLIPSIM}(I, \mathcal{P}_{\text{text}}) = \sum_l \frac{\left\langle \textrm{CLIP}(I_l), \textrm{CLIP}(\mathcal{P}_{\text{text}}) \right\rangle}{\left| \textrm{CLIP}(I_l) \right| \left| \textrm{CLIP}(\mathcal{P}_{\text{text}}) \right|}.
\end{equation}

Following OmniPhysGS~\cite{lin2025omniphysgs}, we use the \texttt{ViT-L/14} variant of \textrm{CLIP}~\cite{radford2021learning} for computing CLIPSIM.

To quantify motion realism, we use the \textit{Energy-Constrained Motion Score} (ECMS) following PhysFlow~\cite{liu2025physflow} and defined as
\[
    \textrm{ECMS} = \frac{1}{\sum_l \lVert F_{l,l+1} \rVert} + \sum_l \lVert F_{l,l+1} - F_{l-1,l} \rVert^2 + \lVert \nabla^2 F_{l,l+1} \rVert^2.
\]
where $F_{l,l+1}$ is the optical flow between frames $l$ and $l+1$. We estimate $F_{l,l+1}$ directly from the generated video $I$ using RAFT~\cite{teed2020raft} with the \texttt{raft-sintel} set of pretrained weights.

We acknowledge that, although OC, CLIPSIM, and ECMS provide useful quantitative insights, they remain imperfect and can diverge from human judgments.

OC relies on ViCLIP, a learned video–text alignment model trained on caption–video paired datasets that typically lack diverse dynamic behaviors across different materials. Consequently, it assigns close scores to videos of distinct materials when appearance, scene layout, geometry, lighting, environment, and force conditions are identical (see Tab.~2 in the main paper).

CLIPSIM evaluates each frame independently against the prompt and therefore cannot assess whether the temporal dynamics reflect the specified material properties. For example, in the Droplet scene (Figure \ref{fig:metric-examples}, Top), PhysDreamer \cite{zhang2024physdreamer} achieves the highest CLIPSIM score despite modeling the droplet as an elastic object rather than liquid, because each frame visually resembles a water droplet, CLIPSIM overlooks the incorrect temporal behavior.

ECMS, in contrast, ignores prompt adherence entirely. By focusing solely on trajectory matching, it may reward unrealistically smooth motions that contradict the intended material response. For instance, OmniPhysGS \cite{lin2025omniphysgs} attains the best ECMS score in Figure \ref{fig:metric-examples} (Bottom), even though the toothpaste neither “collapses” nor the droplet fully “splashes,” resulting in motion that conflicts with the prompt.

Developing improved evaluation metrics is an important direction for future work. One key advancement would be a prompt-adherence metric that captures temporal dynamics under fixed scene and force conditions. Another would be a holistic physical-realism metric that jointly evaluates material properties and dynamic behavior as specified by the text prompt. These enhancements, however, lie beyond the scope of the present study.

\clearpage

\end{document}